\PassOptionsToPackage{unicode}{hyperref}
\PassOptionsToPackage{hyphens}{url}
\documentclass[]{article}
\usepackage{caption}
\usepackage{longtable}
\usepackage{booktabs}
\usepackage[export]{adjustbox} % Add this to your preamble
\usepackage[utf8]{inputenc}
\usepackage[T1]{fontenc}

\usepackage{authblk} % Essential for multiple affiliations

\usepackage{hyperref} % Allows clickable emails

\title{Evaluating Health Misinformation in Low-Resource Languages: Integrating Small Language Models with a Culturally-Sensitive Responsible NLP Framework (Bangla as a Case Study)}

\date{\today} % \today automatically generates current date
\usepackage{xcolor}
\usepackage{amsmath,amssymb}
\usepackage{booktabs}
\usepackage{comment}
\usepackage{tabularx}
\usepackage{iftex}
\usepackage{pdflscape}
\usepackage{booktabs}
\ifPDFTeX
  \usepackage[T1]{fontenc}
  \usepackage[utf8]{inputenc}
  \usepackage{textcomp} % provide euro and other symbols
\else % if luatex or xetex
  \usepackage{unicode-math} % this also loads fontspec
  \defaultfontfeatures{Scale=MatchLowercase}
  \defaultfontfeatures[\rmfamily]{Ligatures=TeX,Scale=1}
\fi
\usepackage{lmodern}
\ifPDFTeX\else
\fi
\IfFileExists{upquote.sty}{\usepackage{upquote}}{}
\IfFileExists{microtype.sty}{% use microtype if available
  \usepackage[]{microtype}
  \UseMicrotypeSet[protrusion]{basicmath} % disable protrusion for tt fonts
}{}
\makeatletter
\@ifundefined{KOMAClassName}{% if non-KOMA class
  \IfFileExists{parskip.sty}{%
    \usepackage{parskip}
  }{% else
    \setlength{\parindent}{0pt}
    \setlength{\parskip}{6pt plus 2pt minus 1pt}}
}{% if KOMA class
  \KOMAoptions{parskip=half}}
\makeatother
\usepackage{longtable,booktabs,array}
\usepackage{multirow}
\usepackage{calc} % for calculating minipage widths
\usepackage{etoolbox}
\makeatletter
\patchcmd\longtable{\par}{\if@noskipsec\mbox{}\fi\par}{}{}
\makeatother
\IfFileExists{footnotehyper.sty}{\usepackage{footnotehyper}}{\usepackage{footnote}}
\makesavenoteenv{longtable}
\usepackage{graphicx}
\usepackage{float}
\makeatletter
\newsavebox\pandoc@box
\newcommand*\pandocbounded[1]{% scales image to fit in text height/width
  \sbox\pandoc@box{#1}%
  \Gscale@div\@tempa{\textheight}{\dimexpr\ht\pandoc@box+\dp\pandoc@box\relax}%
  \Gscale@div\@tempb{\linewidth}{\wd\pandoc@box}%
  \ifdim\@tempb\p@<\@tempa\p@\let\@tempa\@tempb\fi% select the smaller of both
  \ifdim\@tempa\p@<\p@\scalebox{\@tempa}{\usebox\pandoc@box}%
  \else\usebox{\pandoc@box}%
  \fi%
}
\def\fps@figure{htbp}
\makeatother
\ifLuaTeX
  \usepackage{luacolor}
  \usepackage[soul]{lua-ul}
\else
  \usepackage{soul}
\fi
\usepackage{bookmark}
\IfFileExists{xurl.sty}{\usepackage{xurl}}{} % add URL line breaks if available
\hypersetup{
  hidelinks,
  pdfcreator={LaTeX via pandoc}}
\author{}
\date{}

\begin{document}
\maketitle
\vspace{-2cm}
\author{
    \begin{tabular}{c}
        Farnaz Farid\textsuperscript{1}, 
        Raihan Alam\textsuperscript{2}, 
        Al Al-Areqi\textsuperscript{1}, 
        Farhad Ahamed\textsuperscript{1}, \\ 
        Muhammad Hassan Khan\textsuperscript{3,4}, 
        Sadia Hossain\textsuperscript{5}, 
        Irena Veljanova\textsuperscript{1,6}, 
        Anika Tabassum Binte Hossain\textsuperscript{7}
    \end{tabular}
}

\begin{center}
    \small
    \textsuperscript{1}School of Social Sciences, Western Sydney University; 
    \textsuperscript{2}Microsoft; 
    \textsuperscript{3}School of Computer, Data and Mathematical Sciences, Western Sydney University; \\
    \textsuperscript{4}School of Business, Western Sydney University; 
    \textsuperscript{5}School of Medicine, Western Sydney University; 
    \textsuperscript{6}Excelsia College; 
    \textsuperscript{7}Faulconbridge Health Centre
\end{center}
\begin{center}
    \footnotesize
    \texttt{\{farnaz.farid, A.Al-areqi, farhad.ahamed, hassan.khan, sadia.hossain\}@westernsydney.edu.au} \\
    \texttt{raihanalam@microsoft.com, irena.veljanova@excelsia.edu.au, atb.hossain@gmail.com}
\end{center}

\vspace{1cm} % Adds space before the abstract starts
\textbf{Abstract}

The emergence of Artificial Intelligence (AI) and adaptive chatbots, and their uses, have increased significantly across every transaction, ranging from simple information to critical health misinformation. Such rapid proliferation is resulting in escalating threats to digital environments, causing global concerns. Digital health services have expanded access to health information, making it accessible with a single click. This accessibility is causing critical challenges by enabling malicious actors to infect and spread misinformation through novel deceptive practices. Examples include fabricating social media endorsements from prominent medical professionals. AI technologies, while serving as a foundational enabler of modern social media and digital health services, exert a bivalent effect by simultaneously acting as both a combatant against misinformation and a vector for its spread. A prevalent challenge in mitigating this issue arises in non-English contexts and low socioeconomic classes, where limited data hinders the training of AI models for effective detection. Consequently, culturally and linguistically diverse (CALD) communities struggle to access trustworthy health information through AI-driven tools. Current AI tools underperform due to a lack of training data and are largely unable to consider language nuances and traditions in non-English contexts. This research addresses these gaps by proposing a foundation for designing culturally and linguistically friendly AI-based health misinformation detectors and providing a dashboard for medical professionals to analyze this misinformation, a critical step toward mitigating a growing concern among CALD populations. To this end, we conduct a series of experiments using a Bangla-translated health misinformation dataset to evaluate the performance of various Small Language Models (SLMs). SLMs are particularly relevant in this context given the frequent underperformance of Large Language Models (LLMs), which often stems from insufficient domain-specific knowledge and the prohibitive costs of resource-intensive fine-tuning. The results demonstrate that Phi-4 is the superior model, achieving an ideal balance between precision and recall in claim extraction. Then, to mitigate the limitations of SLMs, we design and test a novel health misinformation detection framework grounded in Responsible Natural Language Processing (NLP). This framework moves beyond purely technical metrics to incorporate cultural sensitivity, potential for harm, and communication quality, thereby providing a holistic lens for evaluating misinformation in low-resource languages. Ultimately, this work offers a foundational approach for developing future models that prioritize both accuracy and cultural responsiveness within CALD communities.


\section{Introduction}

Multimodal misinformation now spreads across multiple modalities, such as both structured and unstructured datasets. This includes textual, video, and audio formats, which have proliferated in every aspect of life in one way or another. Fake news related to politics, elections, health, and conflicts gained significant momentum, and people are increasingly relying on social
media for information. This information is mostly collected or generated without proper authenticity, hence leading to and affecting societies and individuals. The rise of AI technologies and the adoption of
Large Language Models (LLMs) are playing a key role in today's digital
ecosystems in generating sophisticated and credible misinformation
{[}1{]}. According to NewsGuard {[}2{]}, the tendency for the top ten
leading chatbots to repeat false information has almost doubled, from
18\% to 35\%, in a year. Table 1 shows some of the data from the
NewsGuard report.

\textbf{Table 1.} Percentage of News Topics that Spread False Claims

{\def\LTcaptype{none} % do not increment counter
\begin{longtable}[]{@{}
  >{\raggedright\arraybackslash}p{(\linewidth - 2\tabcolsep) * \real{0.1514}}
  >{\raggedright\arraybackslash}p{(\linewidth - 2\tabcolsep) * \real{0.2539}}@{}}
\toprule\noalign{}
\begin{minipage}[b]{\linewidth}\raggedright
\textbf{Chatbot}
\end{minipage} & \begin{minipage}[b]{\linewidth}\raggedright
\textbf{Fail rate (\%)}
\end{minipage} \\
\midrule\noalign{}
\endhead
\bottomrule\noalign{}
\endlastfoot
Claude & 10 \\
Gemini & 16.67 \\
Grok & 33.33 \\
You.com & 33.33 \\
Mistral & 36.67 \\
Copilot & 36.67 \\
ChatGPT & 40 \\
Meta & 40 \\
Perplexity & 46.67 \\
Inflection & 56.67 \\
\end{longtable}
}

The World Economic Forum, in its Global Risk report {[}3{]}, has
identified generative AI-fueled information powered by LLM as a leading short-term
threat to global stability and the democratic process. Due to the emergence and accessibility of technology, young people primarily rely on social media for health-related information. Furthermore, social media
plays a crucial role in disseminating and propagating misinformation due to its availability in all parts of the
world's global reach {[} 4 {]}. Among all such misinformation, health
misinformation poses a significant threat to society and can no longer be
viewed merely as an academic nuisance {[}5{]}. Like other domains, the
emergence of generative AI (genAI) and deepfakes has further fueled
health misinformation {[}6{]}. Examples include scammers using deepfake
images of medical professionals to sell products or manipulating
legitimate videos to falsely endorse items, which have garnered millions
of views. For example, drawing from mental health perspectives, a recent
news report from the Guardian reveals that people are relying more on
social media than ever for mental health support, which is seen as an
opportunity by influencers to spread misinformation, involving misused
therapeutic language, ``quick fix'' solutions and false claims {[}7{]}.
The proliferation of online platforms and the volume of user-generated content have made systematic verification of claims increasingly difficult and thus have become a major global challenge in our digital era. These can span from vaccine
hesitancy to the stigmatization of evidence-based pain medications like
paracetamol. This has far-reaching consequences for population health,
affecting trust in institutions, and the uptake of evidence-based care.

Another worrying fact is that health misinformation is not evenly
distributed, with some populations more affected than others. For
example, COVID-19 vaccine hesitancy was most apparent in migrant and
refugee communities in Australian suburbs {[}28{]}. This relates to another critical aspect of health misinformation, namely that Culturally
and Linguistically Diverse (CALD) populations may face a heightened risk
of exposure to health misinformation due to structural and informational
barriers, such as limited access to culturally or linguistically
appropriate health resources and reliance on informal or social media
networks for health information {[}8,9{]}. Existing public health
communication in Australia often lacks cultural nuance and linguistic
accessibility {[}29{]}. Despite the rapid global rise of
Generative-AI-fueled health misinformation, current detection models
often exhibit reduced accuracy in culturally complex or non-English
contexts {[}12,13,14{]}. Their performance in non-English settings
remains underexplored, and the major challenge is the scarcity of
reliable datasets in low-resource languages such as Arabic, Chinese,
Turkish, Bangla, Vietnamese and many more {[}15-18{]}, where no
well-established health misinformation datasets currently exist. While
these factors highlight the human and contextual dimensions of
vulnerability, technological limitations also contribute to
susceptibility to misinformation. For instance, Large Language Models
(LLMs) may inadvertently generate or amplify inaccurate health content
due to biases in training data, limited representation of non-English
languages {[}10{]}, or contextual misinterpretation---thereby
reinforcing the spread of misinformation among non-English-speaking
audiences {[}11{]}.

LLMs, as double-edged technologies, play a dual role in both generating
and detecting misinformation. Furthermore, LLMs have been reported to
underperform in specialized domains such as healthcare and law due to
insufficient domain-specific knowledge and the need for fine-tuning,
which is cost- and resource-intensive {[}30{]}. As a result, Small
Language Models (SLMs) have been seen as an alternative to LLMs. SLMs
have been tested for low-resource languages with promising results
{[}31,32{]}. However, despite such research progress, no expert has yet
annotated a widely accepted health misinformation dataset in Bangla,
which creates a barrier to building and deploying AI-driven
misinformation detection tools.

To further explore their potential in real-world low-resource settings,
this paper makes three key contributions:

\begin{enumerate}
\def\labelenumi{\arabic{enumi}.}
\item
  We create the first expert-validated Bangla health misinformation
  dataset by translating MHMisinfo: Video-based Mental Health
  Misinformation {[}19{]}, initially in English. The translated dataset
  has been reviewed and validated by native-speaking domain experts to
  ensure linguistic and contextual accuracy. The health domain expert
  has also annotated the health claims and health misinformation in the
  translated dataset.
\item
  We evaluate a suite of SLMs on Bangla health misinformation detection
  and compare the results with the human expert-annotated ground truth
  dataset.
\item
  We analyze the cross-lingual performance degradation and highlight the
  implications of CALD populations relying on AI systems that operate in low-resource languages.
  \item
  We first design and test a multi-dimensional framework, which was grounded in a responsible natural language processing (NLP) model that goes beyond technical metrics to incorporate cultural sensitivity, hard potential, and communication quality dimensions, particularly critical for evaluating health misinformation in non-English-language contexts.

\end{enumerate}
To this end, the rest of the paper has been organized as follows below: Section 2 discusses the available literature from conceptual, technical, and socio‑cultural perspectives that inform this work and Section 3 presents the methodology with detailed data analysis and experimented SLM models. Section 4 illustrates the results, followed by a thorough discussion and analysis outlining limitations of this work in Section 5. Section 6 outlines a framework for evaluating health misinformation in non-English contexts, and Section 7 presents concluding remarks and future work.
\section{Literature Review}

\subsection{Background and Motivation}

This section draws on existing research to establish the conceptual, technical, and socio‑cultural gaps that motivate the conduct of this study. It begins
by examining the role of different Large Language Models (LLMs) in generating and
amplifying misinformation among individuals of low socioeconomic class, followed by a review of their well‑documented
limitations in low‑resource settings, particularly in non‑English
languages. Then it discusses the advantages of SLMs as a more practical
alternative. Building on this, the section then highlights how these
linguistic and technical constraints intersect with the unique
vulnerabilities faced by Culturally and Linguistically Diverse (CALD)
communities, where structural, social, and technological barriers compound the risks associated with health misinformation. Together, these strands of literature underscore the need for more culturally grounded, language‑specific approaches, particularly for languages such as Bangla, where no health‑misinformation datasets currently exist.

\subsection{LLMs and Misinformation}

Generative AI increases the threat of misinformation by enabling the credible creation of compelling content at large scale {[}1{]}. The data suggest that the general rate of
disinformation generated by top chatbots has doubled over the past year
{[}2{]}. LLMs can generate personalized misinformation, such as false
news headlines tailored to users' demographic profiles, which are particularly difficult for individuals to detect {[}20{]}. Risks specific to the health domain include LLMs\textquotesingle{} capacity to
generate extensive misinformation blogs about cancer, blood sugar and comprehend with fabricated citations and demographic targeting, and by demonstrating scalable,
tailored health disinformation {[}21{]}. Furthermore, LLMs tend to
generate inaccurate content, or "hallucinations," which are significantly more likely to occur in politically and culturally sensitive domains compared to purely scientific or health topics
{[}22{]}. Even when generating factually accurate content, a distinct
risk arises from overgeneralized health messaging, which is misleading
because it is limited in applicability or relevance to a broader
audience. Health Expert Content Creators (HECCs) noted that the adverse
consequences of this generalized messaging can be as severe as outright
misinformation {[}23{]}.

\subsection{Limitations of LLMs in Low-Resource and Non-English Languages}

Another critical limitation is the Western-centric bias in current LLM
research. The data shows that most studies originating in North America and Europe,
resulting in a limited understanding of how non-English-speaking
populations use social media as a source of information and how LLMs affect their health {[}1{]}.
Yet despite training on large multilingual datasets, open-source LLMs
demonstrate limited effectiveness in languages such as Vietnamese
{[}17{]}. The study in {[}15{]}, which focuses on GPT-4 across Arabic
dialects, demonstrates that dialectal variations pose a unique challenge
for LLMs. One proposed workaround is Cross-lingual transfer learning (training on a high-resource
language and testing on a low-resource language). This is explored as a
potential solution for languages that lack sufficient human-annotated
datasets for misinformation detection {[}16{]}. Studies comparing
transfer across languages (including English, Arabic, Chinese, Turkish,
and Polish) reveal that transfer success hinges more on
domain/contextual similarity than on linguistic similarity. For example,
good transfer performance was observed between Arabic and Chinese, due
to their shared focus on COVID-19-related misinformation.

In contrast, transfer between the linguistically related English and
Polish was poor, largely due to contextual differences. The multilingual
mDeBERTa model was found to be effective for this task transfer. LLMs
possess strong semantic understanding and can be leveraged for detection
{[}19{]}. For Mental Health Misinformation (MHMisinfo) detection on
platforms like YouTube, few-shot in-context learning (ICL) with LLMs
(e.g., GPT-4 and Mistral-0.1) proved effective, often outperforming
gradient-based models.

A further challenge is the scarcity of labelled datasets in
low-resource languages for training LLM-based detection models.
For instance, the study in {[}41{]}
examines 43 health misinformation detection papers and reports that 72\%
studied English-only corpora, with only 2 papers investigating
multilingual or cross-lingual detection, which indicates a critical limitation in the existing literature.
\subsection{Small Language Models (SLMs)}

Fine-tuning large language models (LLMs) require significant
computational power, specialized infrastructure, and high financial
cost. These constraints make it difficult for practitioners and
moderators to adapt LLMs for domain-specific or community-level content
moderation tasks. In response to these limitations, recent research has
increasingly explored the use of small language models (SLMs) as a more
practical alternative. SLMs offer a lighter and more cost-effective
approach, while still maintaining strong performance across a range of
natural language understanding tasks. Studies suggest that well-tuned
SLMs can match or even exceed the performance of larger models in
content moderation settings {[}36{]}.

Khan et al. (2021) {[}37{]} conducted a large-scale benchmark study
comparing traditional machine learning methods, deep learning models,
and SLMs using a dataset of nearly 80,000 real and fake news articles
across multiple topics. Their results showed that SLMs consistently
outperformed both traditional classifiers and deep neural networks.
Among the models tested, RoBERTa achieved the strongest results,
outperforming BERT, DistilBERT, ELECTRA, and ELMo, with accuracy,
precision, recall, and F1 scores reaching 0.96. Importantly, the study
also demonstrated that SLMs retain strong performance even when trained
on relatively smaller datasets, highlighting a key advantage over more
data-hungry deep learning models.

Earlier comparative studies further support this progression in
misinformation detection techniques. The study in {[}38{]} evaluated
traditional classifiers alongside ensemble methods, including bagging,
boosting, and voting, across three datasets. Their findings indicated
that bagging and boosting ensembles consistently outperformed voting
classifiers and standalone traditional models. These studies highlight a
clear shift toward language-model-based approaches for misinformation
detection. Recent evidence suggests that SLMs, in particular, strike a
favorable balance between computational efficiency and detection
performance, making them well-suited for scalable and domain-specific
content moderation applications, including the detection of
health-related misinformation.

\subsection{CALD Challenges in Health Misinformation}

CALD communities face substantial language, resource, and technical
barriers in terms of socioeconomic contexts, when it comes to health
misinformation and more specifically, AI-related health
misinformation. Table 2 details the challenges of health
misinformation in the CALD context.

\begin{quote}
\textbf{Table 2.} Health Misinformation in CALD Context
\end{quote}

{\def\LTcaptype{none} % do not increment counter
\begin{longtable}[]{@{}
  >{\raggedright\arraybackslash}p{(\linewidth - 4\tabcolsep) * \real{0.1825}}
  >{\raggedright\arraybackslash}p{(\linewidth - 4\tabcolsep) * \real{0.3228}}
  >{\raggedright\arraybackslash}p{(\linewidth - 4\tabcolsep) * \real{0.4651}}@{}}
\toprule\noalign{}
\begin{minipage}[b]{\linewidth}\raggedright
\textbf{Challenge Area}
\end{minipage} & \begin{minipage}[b]{\linewidth}\raggedright
\textbf{Details}
\end{minipage} & \begin{minipage}[b]{\linewidth}\raggedright
\textbf{Impact of Health Misinformation}
\end{minipage} \\
\midrule\noalign{}
\endhead
\bottomrule\noalign{}
\endlastfoot
Language and data imbalance & LLMs lack sufficient, high-quality
training data for most non-English languages and dialects. & LLMs fail
to reliably identify false claims, leading to \textbf{inconsistent or
inaccurate health advice} when a user asks a question in their native
language {[}15{]}. \\
Lack of cultural sensitivity & LLMs often lack understanding of
culture-specific health beliefs, practices (e.g., traditional medicine),
and the unique social stigma surrounding certain conditions (e.g.,
mental health). & A response might be factually correct in a
Western/English context but \textbf{culturally inappropriate or
confusing} for a CALD user, leading them to reject the correct advice
and seek information from unreliable sources {[}15,24-26{]}. \\
Structural and information barriers & CALD populations often experience
structural and informational barriers, such as limited access to
culturally or linguistically appropriate health resources. & These
communities often rely on platforms like WhatsApp or WeChat, as well as
social media, which serve as \textbf{breeding grounds for
misinformation}. When they seek clarification from an LLM, the model's
inconsistent performance in their language provides no reliable
safeguard {[}24{]}. \\
Lack of trust in authority & Historical experiences or current events
can lead to low trust in government or official health systems among
migrant and refugee groups. & When official public health messaging is
mistrusted, LLMs that accurately reflect official information may be
viewed with similar skepticism, making the user more susceptible to
\textbf{misinformation spread through community networks} or non-English
media {[}9,27{]}. \\
\end{longtable}
}

These challenges highlight the critcal gaps To address these gaps, our
work aims to use SLMs to address some of LLMs\textquotesingle{}
shortcomings. SLMs have been used in low-resource language contexts, as
evidenced in {[}31,32{]}. The popularity of SLMs in domain-specific
applications is also gaining momentum in healthcare, science, finance,
law, and many other fields {[}33, 34, 35{]}. The aim of our work is as
follows:

\begin{itemize}
\item
  Develop new evaluation datasets or benchmarks for health
  misinformation focused on CALD languages beyond the few existing
  examples, such as Arabic, Chinese, Polish, Turkish, and Vietnamese.
\item
  Integrate culturally specific health misconceptions and linguistic
  nuances (e.g., idioms, colloquialisms, or religious context, as hinted
  by MHMisinfo findings) into cross-lingual detection models.
\item
  Characterize the specific types and linguistic features of
  AI-generated health misinformation (including overgeneralized
  messaging) as they manifest in CALD or low-resource languages.
\item
  Investigate whether the performance variance observed in languages
  such as Bangla (e.g., susceptibility to prompts, toxicity) is linked
  to specific health topics and cultural sensitivities, requiring unique
  detection strategies.
\item
  Empirically evaluate and compare different SLMs for detecting health
  misinformation across a high-resource source (e.g., English) and a
  low-resource target language, focusing on improving the "recovery
  ratio" observed in existing work.
\item
  Determine the optimal data augmentation or fine-tuning techniques
  necessary to achieve reliable detection performance in low-resource
  languages using SLMs.
\item
  Assess the durability and consistency of detection and mitigation
  strategies (e.g., prompt engineering, content moderation) when applied
  to health misinformation generated in CALD languages over extended
  periods.

  In this work, we report the early results of the accomplishments
  mentioned above.
\end{itemize}

\section{Methodology}

The section outlines the methodology employed and used in this research. This includes
encompassing dataset creation, annotation, and experimental evaluation.
The methodology pipeline contains five major stages: 1)dataset
construction and preparation, 2) ground-truth annotation, 3) SLM‑based claim
extraction, 4) misinformation detection, 5) prompt engineering, and 6)
model‑ground‑truth comparison via evaluation pipelines. Figure 1 below
illustrates the comprehensive workflow.
\begin{figure}
\centering
\includegraphics[width=0.8\textwidth]{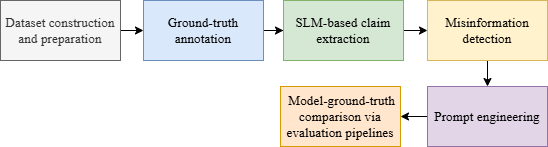}
\caption{Workflow of this work}
\label{fig:Fig. 1}
\end{figure}

\subsection{Dataset Construction and Preparation}

The dataset construction and preparation stage involved three core
milestones, starting from 1) data cleaning (raw data), 2) data transformation into a usable format, and 3) expert
validation.

\subsubsection{Source Dataset}\label{source-dataset}

The dataset used in this research was primarily sourced from {[}19{]}. The dataset
comprises of 739 videos (639 from YouTube and 100 from Bitchute) and
135,372 comments, thus utilising an expert-driven annotation schema. The raw data was first filtered and cleaned as part of the initial process.
In that process, transcriptions shorter than 100 words were
removed, along with several unnecessary columns. Following this filtering process, 286 videos were retained for use in the experimental evaluation.

\textbf{Table 3} Source Dataset Description

{\def\LTcaptype{none} % do not increment counter
\begin{longtable}[]{@{}
  >{\raggedright\arraybackslash}p{(\linewidth - 2\tabcolsep) * \real{0.2246}}
  >{\raggedright\arraybackslash}p{(\linewidth - 2\tabcolsep) * \real{0.7075}}@{}}
\toprule\noalign{}
\begin{minipage}[b]{\linewidth}\raggedright
\textbf{Column Name}
\end{minipage} & \begin{minipage}[b]{\linewidth}\raggedright
\textbf{Description}
\end{minipage} \\
\midrule\noalign{}
\endhead
\bottomrule\noalign{}
\endlastfoot
video\_view\_count & View count of the video, at the time of data
collection \\
video\_like\_count & Like count of the video, at the time of data
collection \\
video\_comment\_count & Comment count of the video, at the time of data
collection \\
label & Overall mental health misinformation label of the video. 0 =
non-MHMisinfo videos, and -1 = MHMisinfo videos \\
label\_ioi & "Information of Interventions" label of video, annotated by
experts. 1 = High-quality information on interventions, -1 = Low-quality
information on interventions \\
label\_ebt & "Evidence-based Treatment" label of video, annotated by
experts. 1 = Encourages evidence-based treatment, -1 = Discourages
evidence-based treatment \\
label\_aoc & "Alignment of Consensus" label of video, annotated by
experts, 1 = High Alignment with Consensus, -1 = Low Alignment with
Consensus \\
platform & Platform of the video \\
\end{longtable}
}

%\textbf{Table 3} Source Dataset Description

\subsubsection{Translation Procedure}\label{translation-procedure}

%remove the quote as this is normal text-Hassan

As stated before, there's no credible Bangla misinformation dataset. As
a result, the cleaned MHMisinfo: Video-based Mental Health Misinformation dataset is then translated into Bangla. To translate, we used a hybrid human-in-the-loop framework: we translated the dataset using Azure translation services and then asked two native Bangla‑speaking experts to check each segment for linguistic fidelity
and cultural appropriateness. Among the two experts, one was the domain
expert who also checked for medical correctness of translation. This
step addressed known issues where machine translations struggle with
health terminology and domain-specific nuance in low‑resource languages
{[}15--17{]}. Corrections were incorporated iteratively until consensus
was reached. Figure 2 shows this process.

\begin{figure}
\centering
\includegraphics[width=0.8\textwidth]{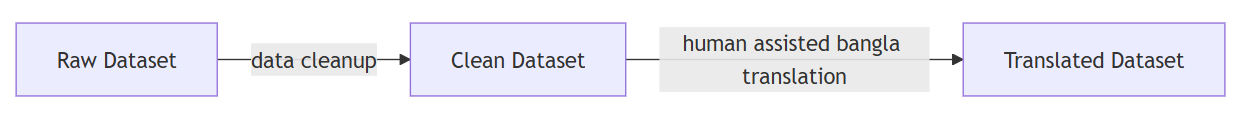}
\caption{Dataset Translation}
\label{fig:Fig.2}
\end{figure}

\subsubsection{Expert Validation}\label{expert-validation}

%remove the quote as this is not quote. this is normal para- Hassan
After the translation is done, the native domain expert checked the
appropriateness of the translation, based on grammatical accuracy,
semantic preservation, contextual appropriateness and correct medical
meaning
% can you pls complete above para?

\subsection{Ground-Truth Annotation}

The next step in this research is to create a ground-truth dataset
annotated by a human expert to extract health-related claims and their
accuracy. This is illustrated in Figure 3.
\begin{figure}
\centering
\includegraphics[width=0.8\textwidth]{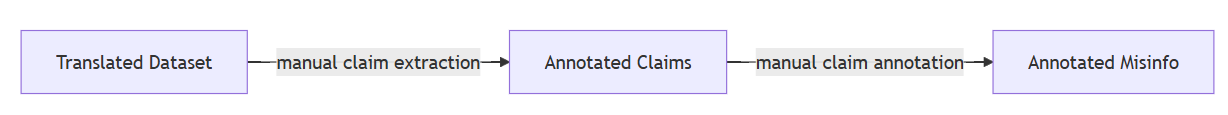}
\caption{Dataset Annotation by Human Experts}
    \label{fig:Fig.3}
\end{figure}

%\subsubsection{Claim Extraction by Human
%Experts}\label{claim-extraction-by-human-experts}

%\begin{quote} remove it- Hassan
\begin{comment}
    
The translated dataset is used for manual claim extraction. This step
involves the human expert reviewing the translated content in the CSV
file to identify and isolate specific, discrete statements that function
as health-related claims. This step yields the annotated claims dataset.
%\end{quote}
\end{comment}

\subsubsection{Misinformation Annotation}

The resulting annotated claims dataset is then used for manual claim
annotation. The core function of the human expert in this stage is to
assess the accuracy of the extracted health-related claims. This
involves checking the factual content against reliable health sources to
determine if the claim is factual or constitutes misinformation. The
culmination of this manual annotation process is the annotated misinfo
dataset. This final dataset contains the verified ground-truth labels
indicating which claims in the Bangla text are misinformation. The
overall purpose of involving a human expert in annotating the cleaned
dataset is to extract health-related claims and assign their
corresponding accuracy ratings. This ensures that the labels used to
evaluate the LLMs reflect true linguistic and contextual nuances
relevant to health information in Bangla.

\subsection{SLM-Based Claim Extraction and Misinformation Detection}

In this step, we set up the experimental environment to evaluate
misinformation detection using SLM models. We test a range of models,
which are described in the subsequent sections. The evaluation was conducted on the 286 videos retained
after the filtering process, encompassing their associated
Bangla-translated health claims.
% added about vidoes used
\subsubsection{Models Evaluated}
% added subsubsection-Hassan
The evaluated SLMs are described in Table 4.
\setcounter{table}{3}
\begin{longtable}{p{0.25\textwidth} p{0.15\textwidth} p{0.12\textwidth} p{0.40\textwidth}}
\caption{Classification of Evaluated Models as Small Language Models (SLMs)} \\
\toprule
\textbf{Model} & \textbf{Parameter Size} & \textbf{Category} & \textbf{Notes} \\
\midrule
\endfirsthead

\toprule
\textbf{Model} & \textbf{Parameter Size} & \textbf{Category} & \textbf{Notes} \\
\midrule
\endhead

\bottomrule
\endlastfoot

Phi‑4 (14B) & 14B & SLM & Mid‑sized model optimized for efficient reasoning and long‑context processing. \\
Qwen3-4B & 3.4B & SLM & Compact multilingual model with strong cross‑lingual generalization. \\
Qwen3-8B & 3.8B & SLM & Small‑scale model designed for efficient inference across diverse tasks. \\
Qwen3‑14B & 14B & SLM & Upper‑tier SLM with extended context window and multilingual capability. \\
Llama‑3.1‑8B‑Instruct & 8B & SLM & Lightweight, instruction‑tuned model suitable for local deployment. \\
Gemma‑3‑12B‑IT & 12B & SLM & Instruction‑tuned multimodal model with a large context window. \\
Ministral‑8B‑2512 & 8B & SLM & Efficient 8B‑parameter model using architectural optimizations for rapid inference. \\
\end{longtable}
\subsubsection{3.3.2 Experimental
Environment}\label{experimental-environment}

The eval pipeline takes the SLM-generated file and compares it against
the ground-truth dataset to compute the metrics. The experiment has been
set up in a local environment using Visual Studio Code (a free and
lightweight source-code editor from Microsoft), UV (a fast, all-in-one
Python package), and Git. Figure 4 shows the setup.

\begin{figure}
\centering
\includegraphics[width=0.6\textwidth]{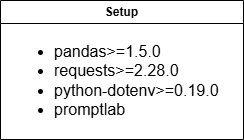}
\caption{Experiment Setup}
\label{Fig:Fig.4}
\end{figure}

\subsection{Prompt Engineering}

We apply a SLM claim processing pipeline to utilize the annotated data
as input prompts. This step is illustrated in Figure 5. We use an SLM to
extract health-related claims from the translated dataset and create an
annotated health claim dataset.

\begin{figure}
\centering
\includegraphics[width=0.8\textwidth]{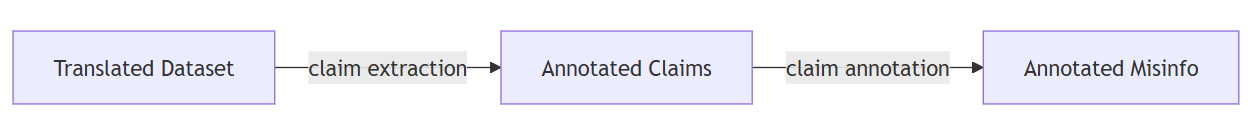}
\caption{SLM claim processing Pipeline}
\label{Fig:Fig.5}
\end{figure}

\subsubsection{3.4.1 Claim Extraction
Prompt}\label{claim-extraction-prompt}

Figure 6 shows an example prompt to extract health claims from the
translated dataset using SLM. Then, SLM prompts are used to extract
health misinformation from those claims, and we annotate the
Misinformation annotated dataset using SLM.

\begin{figure}
\centering
\includegraphics[width=0.8\textwidth,alt={A close-up of a document Description automatically generated}]{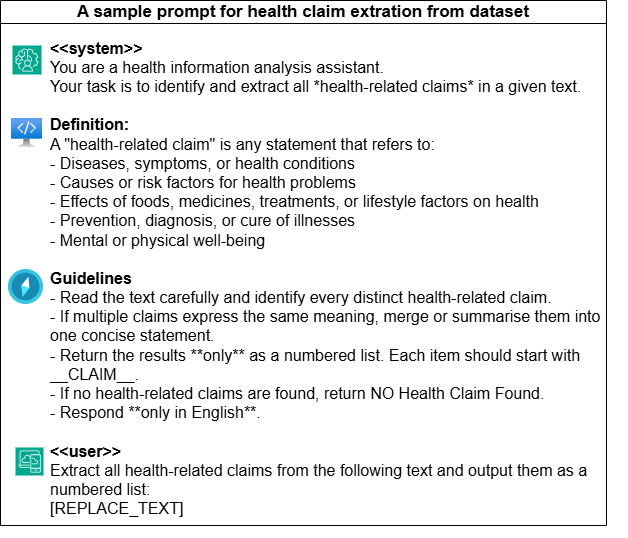}
\caption{Claim extraction using SLM}
\label{fig:Fig. 6}
\end{figure}

\subsubsection{3.4.2 Misinformation Classification
Prompt}\label{misinformation-classification-prompt}

Then, the evaluation pipeline compares the SLM-generated file against
the ground-truth dataset files annotated by human experts to calculate
the metrics. The example prompt is illustrated in Figure 7.

\begin{figure}
\centering
\includegraphics[width=0.8\textwidth,alt={A screenshot of a medical information Description automatically generated}]{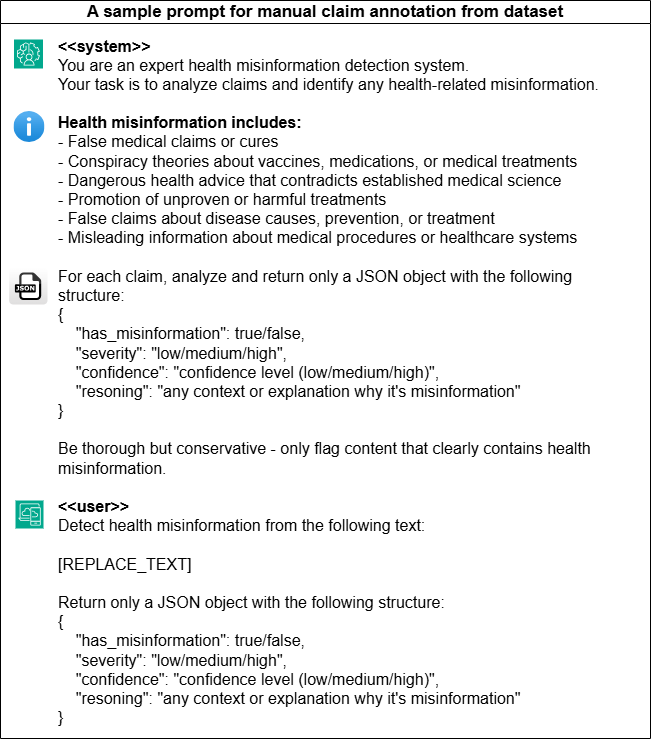}
\caption{Claim annotation using SLM}
\label{fig:Fig.7}
\end{figure}

\subsection{Evaluation Pipeline}

Model outputs are methodically compared against the expert‑annotated
ground truth. We use micro- and macro-averaged metrics to evaluate the
dataset\textquotesingle s performance. Micro metrics are calculated to
ensure overall performance comparison, and macro metrics are calculated
to ensure fairness across videos.

\subsubsection{Micro‑Averaged Metrics}\label{microaveraged-metrics}

Treat all claims across all videos as a single dataset.

\(\text{Precision}_{\text{micro}} = \frac{\sum_{i}^{}TP_{i}}{\sum_{i}^{}TP_{i} + FP_{i})}\)
(1)

\(\text{Recall}_{\text{micro}} = \frac{\sum_{i}^{}TP_{i}}{\sum_{i}^{}TP_{i} + FN_{i})}\)
(2)

\(\text{F1}_{\text{micro}} = 2 \times \frac{{Precision}_{micro}\  \times {\ Recall}_{micro}}{{Precision}_{micro}\  + {\ Recall}_{micro}}\)
(3)

\subsubsection{Macro-Averaged Metrics}

Compute metrics per video, then take the mean across all videos.

\(\text{Precision}_{\text{macro}} = \frac{1}{N}\sum_{i}^{}\text{Precision}_{i}\)
(4)

\(\text{Recall}_{\text{macro}} = \frac{1}{N}\sum_{i}^{}\text{Recall}_{i}\)
(5)

\(\text{F1}_{\text{macro}} = \frac{1}{N}\sum_{i}^{}\text{F1}_{i}\) (6)

Where:

\begin{itemize}
\item
  \textbf{TPᵢ}: True Positives for video \emph{i}
\item
  \textbf{FPᵢ}: False Positives for video \emph{i}
\item
  \textbf{FNᵢ}: False Negatives for video \emph{i}
\item
  \textbf{N}: Number of videos in the dataset
\end{itemize}

\section{Results}

The section presents the performance of all evaluated SLMs on Bangla
health claim extraction and misinformation detection, and it also
compares the results with the original English dataset. Both micro- and
macro-level metrics are reported to provide readers with a comprehensive
overview of accuracy and consistency across all videos. Table 5 shows
the comparison, and the subsequent subsections discuss these results
from different angles.

\footnotesize
\setlength{\tabcolsep}{2pt}
\begingroup
\footnotesize
\setlength{\tabcolsep}{2pt}
\begin{longtable}[]{@{}
  >{\raggedright\arraybackslash}p{(\linewidth - 12\tabcolsep) * \real{0.2200}}
  >{\centering\arraybackslash}p{(\linewidth - 12\tabcolsep) * \real{0.1200}}
  >{\centering\arraybackslash}p{(\linewidth - 12\tabcolsep) * \real{0.1150}}
  >{\centering\arraybackslash}p{(\linewidth - 12\tabcolsep) * \real{0.1100}}
  >{\centering\arraybackslash}p{(\linewidth - 12\tabcolsep) * \real{0.1300}}
  >{\centering\arraybackslash}p{(\linewidth - 12\tabcolsep) * \real{0.1250}}
  >{\centering\arraybackslash}p{(\linewidth - 12\tabcolsep) * \real{0.0930}}@{}}
\caption{English Dataset Results}\tabularnewline
\toprule\noalign{}
\begin{minipage}[b]{\linewidth}\raggedright
Model
\end{minipage} & \begin{minipage}[b]{\linewidth}\raggedright
Prec.\textsubscript{micro}
\end{minipage} & \begin{minipage}[b]{\linewidth}\raggedright
Rec.\textsubscript{micro}
\end{minipage} & \begin{minipage}[b]{\linewidth}\raggedright
F1\textsubscript{micro}
\end{minipage} & \begin{minipage}[b]{\linewidth}\raggedright
Prec.\textsubscript{macro}
\end{minipage} & \begin{minipage}[b]{\linewidth}\raggedright
Rec.\textsubscript{macro}
\end{minipage} & \begin{minipage}[b]{\linewidth}\raggedright
F1\textsubscript{macro}
\end{minipage} \\
\midrule\noalign{}
\endfirsthead
\toprule\noalign{}
\begin{minipage}[b]{\linewidth}\raggedright
Model
\end{minipage} & \begin{minipage}[b]{\linewidth}\raggedright
Prec.\textsubscript{micro}
\end{minipage} & \begin{minipage}[b]{\linewidth}\raggedright
Rec.\textsubscript{micro}
\end{minipage} & \begin{minipage}[b]{\linewidth}\raggedright
F1\textsubscript{micro}
\end{minipage} & \begin{minipage}[b]{\linewidth}\raggedright
Prec.\textsubscript{macro}
\end{minipage} & \begin{minipage}[b]{\linewidth}\raggedright
Rec.\textsubscript{macro}
\end{minipage} & \begin{minipage}[b]{\linewidth}\raggedright
F1\textsubscript{macro}
\end{minipage} \\
\midrule\noalign{}
\endhead
\bottomrule\noalign{}
\endlastfoot
Phi-4 & 65.55 & 31.28 & 42.35 & 9.81 & 18.96 & 12.00 \\
Qwen3-8B & 86.60 & 41.32 & 55.95 & 13.01 & 24.51 & 16.05 \\
Llama-3.1-8B-Instruct & 64.59 & 23.08 & 34.01 & 8.88 & 18.05
& 11.10 \\
Gemma-3-12B-IT & 45.45 & 11.67 & 18.68 & 6.70 & 11.99 & 8.18 \\
Ministral-8B-2512 & 38.76 & 13.39 & 19.9 & 6.24 & 14.00 &
8.27 \\
\end{longtable}
\endgroup
\normalsize

Qwen3-8B achieves the strongest performance across both micro and macro-level metrics,
followed by Phi-4. Llama-3.1-8B-Instruct closely follows Phi-4, while 
Gemma-3-12B-IT ranks second lowest, and Ministral-8B-2512 achieves the weakest overall result.

\begingroup
\footnotesize
\setlength{\tabcolsep}{2pt}
\begin{longtable}[]{@{}
  >{\raggedright\arraybackslash}p{(\linewidth - 12\tabcolsep) * \real{0.2200}}
  >{\centering\arraybackslash}p{(\linewidth - 12\tabcolsep) * \real{0.1200}}
  >{\centering\arraybackslash}p{(\linewidth - 12\tabcolsep) * \real{0.1150}}
  >{\centering\arraybackslash}p{(\linewidth - 12\tabcolsep) * \real{0.1100}}
  >{\centering\arraybackslash}p{(\linewidth - 12\tabcolsep) * \real{0.1300}}
  >{\centering\arraybackslash}p{(\linewidth - 12\tabcolsep) * \real{0.1250}}
  >{\centering\arraybackslash}p{(\linewidth - 12\tabcolsep) * \real{0.0930}}@{}}
\caption{Model comparison results for Bangla Translated
Dataset}\tabularnewline
\toprule\noalign{}
\begin{minipage}[b]{\linewidth}\raggedright
Model
\end{minipage} & \begin{minipage}[b]{\linewidth}\raggedright
Prec.\textsubscript{micro}
\end{minipage} & \begin{minipage}[b]{\linewidth}\raggedright
Rec.\textsubscript{micro}
\end{minipage} & \begin{minipage}[b]{\linewidth}\raggedright
F1\textsubscript{micro}
\end{minipage} & \begin{minipage}[b]{\linewidth}\raggedright
Prec.\textsubscript{macro}
\end{minipage} & \begin{minipage}[b]{\linewidth}\raggedright
Rec.\textsubscript{macro}
\end{minipage} & \begin{minipage}[b]{\linewidth}\raggedright
F1\textsubscript{macro}
\end{minipage} \\
\midrule\noalign{}
\endfirsthead
\toprule\noalign{}
\begin{minipage}[b]{\linewidth}\raggedright
Model
\end{minipage} & \begin{minipage}[b]{\linewidth}\raggedright
Prec.\textsubscript{micro}
\end{minipage} & \begin{minipage}[b]{\linewidth}\raggedright
Rec.\textsubscript{micro}
\end{minipage} & \begin{minipage}[b]{\linewidth}\raggedright
F1\textsubscript{micro}
\end{minipage} & \begin{minipage}[b]{\linewidth}\raggedright
Prec.\textsubscript{macro}
\end{minipage} & \begin{minipage}[b]{\linewidth}\raggedright
Rec.\textsubscript{macro}
\end{minipage} & \begin{minipage}[b]{\linewidth}\raggedright
F1\textsubscript{macro}
\end{minipage} \\
\midrule\noalign{}
\endhead
\bottomrule\noalign{}
\endlastfoot
Phi-4 & 80.86 & 52.65 & 63.77 & 12.7 & 24.9 & 15.49 \\
Qwen3-8B & 85.65 & 41.24 & 55.68 & 12.37 & 25.44 & 15.36 \\
Qwen3-4B & 85.6 & 32.6 & 47.23 & 12.5 & 24.6 & 15.5 \\
Llama-3.1-8B-Instruct & 50.24 & 15.24 & 23.39 & 8.78 & 13.45
& 9.71 \\
Gemma-3-12B-IT & 40.67 & 9.15 & 14.94 & 5.4 & 10.33 & 6.66 \\
Ministral-8B-2512 & 36.36 & 11.13 & 17.04 & 5.45 & 13.22 &
7.23 \\
\end{longtable}
\endgroup
\normalsize

From Table 6, it's evident that across all models, for Bangla translated
dataset Phi-4 achieves the best result with a F1 micro score of 63.8,
followed by qwen3-8B with a F1 score of 55.7 and qwen3-4B (47.2). On the
other hand, Llama‑3.1‑8B, Gemma‑3‑12B, and Ministral‑8B demonstrate
lower performance with a micro F1 score. The results emerge the patterns
that models with stronger multilingual pre‑training (Phi and Qwen
series) outperform models whose training data is less balanced across
non‑English languages (Llama, Gemma, Ministral).

The analysis of the micro-level metrics shows an interesting trend: high
precision but low recall across nearly all models. For example,
Qwen3-8B achieves 85.65\% precision and 41.24 recall, and Qwen3-4B
achieves 85.6 precision and 32.6 recall. This suggests that the models
can accurately predict misinformation; however, they are misclassifying
many positive cases, resulting in many false negatives. This behaviour
is supported by other similar works done in low-resource languages
{[}10{]}, where models struggle to recognize diverse linguistic forms
but remain cautious in their predictions. Phi‑4, although lower in
precision (80.86), achieves significantly higher recall (52.65) than the
Qwen models.

%\begin{quote} this is not quote
Another observation is that the Macro‑averaged metrics reveal a deeper
challenge. Across all models, macro‑F1 scores are extremely low, ranging
from 15.5 (Phi‑4),15.4 (Qwen3-8B), down to 6.7 (Gemma‑3‑12B). Such
results indicate severe inconsistency across videos. This suggests that
the models perform well on some videos but fail almost entirely on
others. Additionally, we observe that micro-averaged recall is
substantially higher than macro-averaged recall. This discrepancy is
expected in imbalanced classification settings. Micro recall aggregates
true positives and false negatives across all samples, thereby giving
greater weight to the majority classes. In contrast, macro recall
computes recall independently for each class and then averages them,
assigning equal importance to both majority and minority classes.

In practice, strong performance on dominant classes can lead to high
micro recall, even when the model performs poorly on underrepresented
classes. In this study, the lower macro recall observed in our results indicates that
the model struggles to effectively capture minority class instances.
This is especially important and critical in the context of misinformation
detection, where minority classes (e.g., misinformation instances) carry the greatest practical significance. Therefore, macro-averaged
metrics offer a more balanced assessment of model performance across
classes when working with an imbalanced dataset like this study. Hence, reporting both micro- and macro-level scores together gives 
a far more honest account of where models succeed and where they fall short.
%\end{quote}

\begin{longtable}[]{@{}
  >{\raggedright\arraybackslash}p{(\linewidth - 2\tabcolsep) * \real{0.2354}}
  >{\raggedright\arraybackslash}p{(\linewidth - 2\tabcolsep) * \real{0.4630}}@{}}
\caption{Comparative Model Insights -}\tabularnewline
\toprule\noalign{}
\begin{minipage}[b]{\linewidth}\raggedright
Model
\end{minipage} & \begin{minipage}[b]{\linewidth}\raggedright
Insights
\end{minipage} \\
\midrule\noalign{}
\endfirsthead
\toprule\noalign{}
\begin{minipage}[b]{\linewidth}\raggedright
Model
\end{minipage} & \begin{minipage}[b]{\linewidth}\raggedright
Insights
\end{minipage} \\
\midrule\noalign{}
\endhead
\bottomrule\noalign{}
\endlastfoot
Phi-4 & \begin{minipage}[t]{\linewidth}\raggedright
\begin{itemize}
\item
  Best overall model with the highest micro-F1 score (63.77) and the
  highest recall (52.65), while maintaining high precision (80.86)
\item
  Provided the most balanced trade-off between identifying relevant
  claims and avoiding false positives
\end{itemize}
\end{minipage} \\
Qwen3-8B \& qwen3-4B & \begin{minipage}[t]{\linewidth}\raggedright
\begin{itemize}
\item
  Highest micro-precision across all models (85.6), indicating strong
  selectivity in claim extraction
\item
  Recall relatively low (41.24\%)
\item
  Indicates conservative extraction behaviour
\item
  Strong multilingual encodings help with Bangla syntax
\end{itemize}
\end{minipage} \\
Llama-3.1-8B-Instruct &
\begin{minipage}[t]{\linewidth}\raggedright
\begin{itemize}
\item
  Moderate precision (50.2) but very poor recall (15.2)
\item
  Suggests limited Bangla representation in pre-training
\item
  Frequently misinterprets conversational Bangla structures
\end{itemize}
\end{minipage} \\
Gemma-3-12B-IT & \begin{minipage}[t]{\linewidth}\raggedright
\begin{itemize}
\item
  Performs weakest overall (F1 = 14.9 micro)
\item
  Very low macro‑F1 (6.66)
\item
  Likely reflects limited tokenization coverage for Bangla
\item
  Indicate both poor aggregate performance and limited consistency
  across videos
\end{itemize}
\end{minipage} \\
Ministral-8B-2512 &
\begin{minipage}[t]{\linewidth}\raggedright
\begin{itemize}
\item
  Similar to Llama and Gemma in poor cross‑lingual adaptation with low
  precision (36.36), low recall (11.13), and low micro-F1 (17.04)
\item
  Tends to over‑ or under‑segment claims
\end{itemize}
\end{minipage} \\
\end{longtable}

To conclude, among the evaluated models, Phi-4 showed the strongest overall claim
extraction performance by achieving the best balance between precision and
recall. After several rounds of testing, the Qwen models were highly precise but seemed to be comparatively less
sensitive, overall indicating that they were more conservative in extracting
claims. In contrast, Llama, Gemma, and Ministral were also tested and showed substantially
weaker recall and lower overall F1 scores, suggesting reduced
effectiveness for Bangla health claim extraction in this dataset.

\section{Discussion}

This section interprets the results and their implications within the context of the research
objectives and existing literature. Moreover, this section highlights the study's implications
and limitations, and suggests future research directions in other low-income societies.

Identifying misinformation with LLMs has proven to be very effective;
for example, fine-tuned GPT-4o and GPT-4o-mini models achieved 98.6\%
accuracy {[}39{]}. However, these models are resource-intensive,
requiring internet connectivity and are therefore not suitable for
stand-alone use. Trained SLMs have a strong future for deployment as
stand-alone models to identify health and other domain-related
misinformation.
Regarding health misinformation, the results show that SLMs are in their early phase of detecting health misinformation.
There is also variation in detection performance across languages,
specifically between English and Bangla. Phi-4 and Qwen models
achieved F1 scores of 63.77\% and 55.68\% respectively. The qwen3-8b model showed consistency in detecting health
misinformation both in English and Bangla language. However, there is
plenty of scope to improve the SLM. SLM are also getting more comparable
with the traditional deep learning models such as CNNs, that achieved
58.6\% accuracy to detect misinformation {[}39{]}.

A notable gap exists between micro- and macro-averaged F1 scores
across all evaluated SLM models. The macro method of F1 scoring varies
significantly from one video to the next. The SLM models have not yet
matured enough to identify various types of health misinformation in the
videos.

The results of this research align with earlier studies' findings, which reported similar issues
in Arabic, Vietnamese, and other low‑resource languages. Past data shows that even when
translation is accurate, non‑English languages carry cultural and
linguistic patterns that multilingual models struggle to capture. The results of this study suggests that Bangla faces the same challenge: general-purpose
LLMs and SLMs simply do not have enough exposure to the full range of
Bangla expressions used in everyday health discussions.

As the F1-macro and F1-micro score varies a lot in different language,
from a broader perspective, it can be concluded that SLM-based health
misinformation detection models will have implications for CALD
populations. Bangla-speaking communities often rely on social media and
family networks for health advice {[}40{]}. If SLM or LLM tools cannot
interpret their language accurately or fairly, misinformation may go
unnoticed or unchallenged. However, the initial performance of Phi‑4 and
the Qwen models compared to the established deep learning CNN model
{[}39{]} suggests that compact SLM models can still serve as useful
foundations for low-resourced devices in community‑focused tools.

\textbf{Limitations}:

There are several limitations in these experiments. In the health
misinformation dataset, translation from English to Bangla was checked
by experts. However, the translation process can still introduce small
shifts in meaning. The video dataset also focuses on multiple health
related misinformation including mental health content, which may behave
differently from other health domains. Another limitation of the
experiments is the lack of fine-tuning of the SLMs into the health
misinformation domain; rather, we relied only on prompting, which
limits performance. Moreover, this dataset comes from two specific
platforms, and patterns may differ on other social channels commonly
used by Bangladeshi or Bangla-speaking CALD communities. These limitations motivate the development of a more holistic
evaluation framework, which is presented in the following section.
%added above - hassan

\section{A Responsible NLP Approach to Evaluating Health Misinformation in Non-English Low-Resource Languages}

The performance variability observed across models and languages in this
study leads to further consideration of new metrics. It is apparent that
standard performance metrics, such as precision, recall, and F1, cannot
capture the nuances of health misinformation in low‑resource languages.
The observation stems from the large difference between Macro- and
Micro-level F1 scores, which range from 63.77\% to 6.66\% across models.
Such large gaps in numbers indicate that there are persistent
inconsistencies in video-level measurement. On the other hand, the high
precision and low recall pattern persists across all models, suggesting
an inability to detect misinformation rather than misclassification.

Since non-English languages are rich in linguistic nuances and the
languages of such communities are closely connected to their medical
practices, relying solely on technical accuracy is unlikely to work.
More metrics across the cultural, community, harm, and risk dimensions
can inform better outcomes for a community oriented toward AI-driven
innovations. As a result, we propose a framework grounded in the
principles of Responsible Natural Language Processing (RNLP) proposed by
{[}42{]}, which advocates for the well-being of the community,
human-centered values, inclusivity and accessibility, and the avoidance
of unfair discrimination against individuals, communities or groups. Our
framework also upholds the integration by engaging human experts to
classify misinformation and assess harm and risk, using human judgment,
and by using human native languages to test different models. By
building on community-centered AI principles to focus more on AI
fairness for marginalized populations, we propose a multi-dimensional
framework, that goes beyond technical metrics to incorporate cultural
sensitivity, potential for harm, and communication quality ---
dimensions particularly critical for evaluating health misinformation in
non-English-language contexts. As depicted in Figure 8, the framework
draws on six dimensions to effectively inform the detection of health
misinformation for non-English texts.

\begin{figure}
\centering
\includegraphics[width=0.8\textwidth]{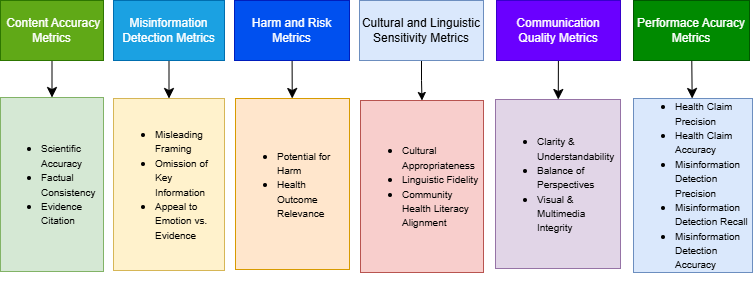}
\caption{The Proposed Framework}
\label{Fig:Fig.8}
\end{figure}

The framework aims to classify health misinformation across six
dimensions. Starting with context accuracy (CA), it moves on to
misinformation detection (M), harm and risk assessment (HR), cultural
and linguistic sensitivity (CLS), communication  quality metrics (CQ), and finally performance accuracy (PA) in the context of AI models.
These lenses are not applied linearly to any misinformation reduction; however, the dimensions are interdependent. For example, linguistic
nuances need to be established before any content accuracy is investigated, and while scientifically validating content accuracy, this should be connected to cultural and communication quality. The harm and risk metrics should eventually be measured to gauge the seriousness of
community impacts and to ensure culturally responsive communication quality, and empirical validation will be performed using performance
accuracy metrics. Table 8 shows how the framework maps to RNLP principles.

\begin{longtable}[]{@{}
  >{\raggedright\arraybackslash}p{(\linewidth - 4\tabcolsep) * \real{0.1934}}
  >{\raggedright\arraybackslash}p{(\linewidth - 4\tabcolsep) * \real{0.1320}}
  >{\raggedright\arraybackslash}p{(\linewidth - 4\tabcolsep) * \real{0.4641}}@{}}
\caption{Mapping of Framework Principles to
RNLP Principles}\tabularnewline
\toprule\noalign{}
\begin{minipage}[b]{\linewidth}\raggedright
\textbf{Dimensions}
\end{minipage} & \begin{minipage}[b]{\linewidth}\raggedright
\textbf{Focus}
\end{minipage} & \begin{minipage}[b]{\linewidth}\raggedright
\textbf{RNLP Principles}
\end{minipage} \\
\midrule\noalign{}
\endfirsthead
\toprule\noalign{}
\begin{minipage}[b]{\linewidth}\raggedright
\textbf{Dimensions}
\end{minipage} & \begin{minipage}[b]{\linewidth}\raggedright
\textbf{Focus}
\end{minipage} & \begin{minipage}[b]{\linewidth}\raggedright
\textbf{RNLP Principles}
\end{minipage} \\
\midrule\noalign{}
\endhead
\bottomrule\noalign{}
\endlastfoot
Content Accuracy (CA) & Is the health claim scientifically grounded? &
Reliability \\
Misinformation Detection (M) & How distorted, misleading and imbalanced
is the content? & Wellbeing and accountability, interrogation \\
Harm and Risk Assessment (HR) & How severe the harm and risk from this
content? & Human-centered values \& Transparency, Privacy and
Security \\
Cultural and Linguistic Sensitivity (CLS) & Is it Culturally valid and
linguistically appropriate for the language? & Fairness \\
Communication quality (CA) & Is content accessible and balanced? &
Accountability and Integration \\
Performance evaluation (PA) & How reliabily the model detects
misinformation? & Reliability, Privacy and Security, interrogation \\
\end{longtable}

\subsection{Content Accuracy Metrics}
The metrics in this category evaluate video content through the lenses of science, facts, and evidence. The process is to check if there is any established evidence from well-established institutions, such as WHO/CDC, peer-reviewed journals, expert consensus, and if the statements made in the video are contradiction-free.
\begin{itemize}
\item
  \textbf{Scientific Accuracy} \textbf{\& Evidence Citation}
\begin{itemize}
\item
  \textbf{Definition:} Degree to which the information in the video is
  supported by established scientific evidence (e.g., peer-reviewed
  research, guidelines from WHO/CDC) and the extent to which the video cites
  reliable evidence or sources.
\item
  \textbf{Evaluation:} Compare claims with trusted sources, such as the presence of references to credible institutions, peer-reviewed
  journals, or expert consensus.
  \end{itemize}

\item
  \textbf{Factual Consistency}
  \begin{itemize}

  \item
    \textbf{Definition:} Whether the facts presented are internally
    consistent and free of contradictions.
\item
  \textbf{Evaluation:} Check for contradictory statements within the video or versus known facts.
  The scale of the metrics is as illustrated in Table 9.
    \end{itemize}
\end{itemize}

\begin{longtable}[]{@{}
  >{\raggedright\arraybackslash}p{(\linewidth - 4\tabcolsep) * \real{0.0933}}
  >{\raggedright\arraybackslash}p{(\linewidth - 4\tabcolsep) * \real{0.2132}}
  >{\raggedright\arraybackslash}p{(\linewidth - 4\tabcolsep) * \real{0.4831}}@{}}
\caption{Content Accuracy
Scale}\tabularnewline
\toprule\noalign{}
\begin{minipage}[b]{\linewidth}\raggedright
\textbf{Scale}
\end{minipage} & \begin{minipage}[b]{\linewidth}\raggedright
\textbf{Label}
\end{minipage} & \begin{minipage}[b]{\linewidth}\raggedright
\textbf{Criteria}
\end{minipage} \\
\midrule\noalign{}
\endfirsthead
\toprule\noalign{}
\begin{minipage}[b]{\linewidth}\raggedright
\textbf{Scale}
\end{minipage} & \begin{minipage}[b]{\linewidth}\raggedright
\textbf{Label}
\end{minipage} & \begin{minipage}[b]{\linewidth}\raggedright
\textbf{Criteria}
\end{minipage} \\
\midrule\noalign{}
\endhead
\bottomrule\noalign{}
\endlastfoot
0 & Inaccurate & Not backed up by any established
literature/institutions \\
1 & Misleading & Accurate content is mixed up with some inaccurate
content out of context to mislead users \\
2 & Partially Accurate & Some evidence available \\
3 & Accurate & Fully backed up by established literature \\
\end{longtable}

\subsection{Misinformation Detection Metrics}

This set of metrics is a combination of misleading framing through
  exaggeration, or selective emphasis a combination of misleading
  framing by using exaggeration, or selective emphasize; some critical
  details are omitted for some benefit to the presenter and usage of
emotional language to draw emotions from the viewers.

\begin{itemize}
\item
  \textbf{Misleading Framing}

  \begin{itemize}
  \item
    \textbf{Definition:} Use of language or presentation style that
    gives a distorted impression (e.g., exaggeration, selective
    emphasis).
  \item
    \textbf{Evaluation:} Rate the extent to which framing may lead to
    misinterpretation of risk/benefits.
  \end{itemize}
\item
  \textbf{Omission of Key Information}

  \begin{itemize}
  \item
    \textbf{Definition:} Whether critical details (e.g., side effects,
    limitations, uncertainties) are missing.
  \item
    \textbf{Evaluation:} Check if risks, alternative treatments, or
    context are ignored.
  \end{itemize}
\item
  \textbf{Appeal to Emotion vs. Evidence}

  \begin{itemize}
  \item
    \textbf{Definition:} Balance between emotional appeals (fear, hope,
    outrage) and factual evidence.

\item
  \textbf{Evaluation:} Measure ratio of emotional language to scientific reasoning.
    \end{itemize}

  The scale of the metrics is as illustrated in Table 10:
\end{itemize}

\begin{longtable}[]{@{}
  >{\raggedright\arraybackslash}p{(\linewidth - 4\tabcolsep) * \real{0.0933}}
  >{\raggedright\arraybackslash}p{(\linewidth - 4\tabcolsep) * \real{0.2132}}
  >{\raggedright\arraybackslash}p{(\linewidth - 4\tabcolsep) * \real{0.4831}}@{}}
\caption{\textbf{Misinformation}
\textbf{Detection} \textbf{Scale}}\tabularnewline
\toprule\noalign{}
\begin{minipage}[b]{\linewidth}\raggedright
\textbf{Scale}
\end{minipage} & \begin{minipage}[b]{\linewidth}\raggedright
\textbf{Label}
\end{minipage} & \begin{minipage}[b]{\linewidth}\raggedright
\textbf{Criteria}
\end{minipage} \\
\midrule\noalign{}
\endfirsthead
\toprule\noalign{}
\begin{minipage}[b]{\linewidth}\raggedright
\textbf{Scale}
\end{minipage} & \begin{minipage}[b]{\linewidth}\raggedright
\textbf{Label}
\end{minipage} & \begin{minipage}[b]{\linewidth}\raggedright
\textbf{Criteria}
\end{minipage} \\
\midrule\noalign{}
\endhead
\bottomrule\noalign{}
\endlastfoot
0 & Severely Misleading Content & Framing may lead to severe misinterpretation
of risk/benefits, and all alternative treatments, contexts are ignored,
treatments, contexts are ignored and emotional language has been used to
severely mislead viewers. \\
1 & Moderate Misleading  & Framing may lead to moderate misinterpretation
of risk/benefits, and many alternative treatments, contexts are ignored,
treatments, contexts are ignored and emotional language has been used to
significantly mislead viewers. \\
2 &  Minor Misleading & Framing may lead to minor misinterpretation of
risk/benefits, and some alternative treatments, contexts are ignored,
treatments, contexts are ignored and emotional language has been used to
slightly mislead viewers. \\
3 & Not Misleading &  The content is fine\\
\end{longtable}

\subsection{Harm and Risk Metrics}

\begin{itemize}
\item
  \textbf{Potential for Harm}

  \begin{itemize}
  \item
    \textbf{Definition:} Likelihood that viewers may adopt unsafe
    behaviors or reject evidence-based treatments due to the video.
  \item
    \textbf{Evaluation:} Classify as low, moderate, or high potential
    harm.
  \end{itemize}
\item
  \textbf{Health Outcome Relevance}

  \begin{itemize}
  \item
    \textbf{Definition:} Whether misinformation relates to severe health
    issues (e.g., cancer treatment, vaccines) vs. less critical topics.

\item
  \textbf{Evaluation:} Weight misinformation severity by its possible real-world health impact.
    \end{itemize}
%\item
%  \textbf{Table} \textbf{11} \textbf{Harm and Risk Scale}
\end{itemize}

%{\def\LTcaptype{none} % do not increment counter
\begin{longtable}[]{@{}
  >{\raggedright\arraybackslash}p{(\linewidth - 4\tabcolsep) * \real{0.0933}}
  >{\raggedright\arraybackslash}p{(\linewidth - 4\tabcolsep) * \real{0.2132}}
  >{\raggedright\arraybackslash}p{(\linewidth - 4\tabcolsep) * \real{0.4831}}@{}}
  \caption{Harm and Risk Scale}\tabularnewline
\toprule\noalign{}
\begin{minipage}[b]{\linewidth}\raggedright
\textbf{Scale}
\end{minipage} & \begin{minipage}[b]{\linewidth}\raggedright
\textbf{Label}
\end{minipage} & \begin{minipage}[b]{\linewidth}\raggedright
\textbf{Criteria}
\end{minipage} \\
\midrule\noalign{}
\endhead
\bottomrule\noalign{}
\endlastfoot
0 & No Harm and Risk involved & No risk of harm and risk. \\
1 & Low Harm and Risk involved & The content may bring slight harm and
risk. \\
2 & Medium Harm and Risk involved & The content may bring medium harm
and risk. \\
3 & High Harm and Risk involved & The content may bring significant harm
and risk. \\
\end{longtable}

\subsection{Cultural and Linguistic Sensitivity Metrics}

\begin{itemize}
\item
  \textbf{Cultural Appropriateness}

  \begin{itemize}
  \item
    \textbf{Definition:} The degree to which health content accounts for
    community-specific beliefs, practices, and values, such as
    traditional medicine, religious health perspectives, or culturally
    specific stigma
  \item
    \textbf{Evaluation:} Expert rating scale from culturally
    inappropriate to fully appropriate
  \end{itemize}
\item
  \textbf{Linguistic Fidelity}

  \begin{itemize}
  \item
    \textbf{Definition:} Whether translated or non-English content
    preserves the original health meaning without introducing ambiguity,
    distortion, or loss of nuance.
  \item
    \textbf{Evaluation:} Back-translation verification or bilingual
    expert review
  \end{itemize}
\item
  \textbf{Community Health Literacy Alignment}

  \begin{itemize}
  \item
    \textbf{Definition:} Whether the complexity and framing of health
    information is accessible to the health literacy level of the target
    CALD community.
  \item
    \textbf{Evaluation:} Readability assessment calibrated to the target
    language and community context
  \end{itemize}
\end{itemize}

\begin{longtable}[]{@{}
  >{\raggedright\arraybackslash}p{(\linewidth - 4\tabcolsep) * \real{0.0933}}
  >{\raggedright\arraybackslash}p{(\linewidth - 4\tabcolsep) * \real{0.2132}}
  >{\raggedright\arraybackslash}p{(\linewidth - 4\tabcolsep) * \real{0.4831}}@{}}
\caption{\textbf{Cultural and Language}
\textbf{Sensitivity} \textbf{Scale}}\tabularnewline
\toprule\noalign{}
\begin{minipage}[b]{\linewidth}\raggedright
\textbf{Scale}
\end{minipage} & \begin{minipage}[b]{\linewidth}\raggedright
\textbf{Label}
\end{minipage} & \begin{minipage}[b]{\linewidth}\raggedright
\textbf{Criteria}
\end{minipage} \\
\midrule\noalign{}
\endfirsthead
\toprule\noalign{}
\begin{minipage}[b]{\linewidth}\raggedright
\textbf{Scale}
\end{minipage} & \begin{minipage}[b]{\linewidth}\raggedright
\textbf{Label}
\end{minipage} & \begin{minipage}[b]{\linewidth}\raggedright
\textbf{Criteria}
\end{minipage} \\
\midrule\noalign{}
\endhead
\bottomrule\noalign{}
\endlastfoot
0 & Inappropriate & Culturally and linguistically inappropriate \\
1 & Culturally and Linguistically Insensitive & Reflects Western or
English-language assumptions that may mislead or alienate CALD users \\
2 & Partially Inappropriate & Some elements of culturally inappropriate
elements. \\
3 & Appropriate & Fully based by cultural and language contexts, \\
\end{longtable}

\subsection{Communication Quality Metrics}

\begin{itemize}
\item
  \textbf{Clarity \& Understandability}

  \begin{itemize}
  \item
    \textbf{Definition:} How clearly health concepts are explained in layperson-friendly language.
  \item
    \textbf{Evaluation:} Apply readability scores (e.g., Flesch-Kincaid) or expert rating.
  \end{itemize}
\item
  \textbf{Balance of Perspectives}

  \begin{itemize}
  \item
    \textbf{Definition:} Whether multiple viewpoints (e.g., medical
    consensus vs. uncertainties) are represented.
  \item
    \textbf{Evaluation:} Rate from one-sided to balanced.
  \end{itemize}
\item
  \textbf{Visual \& Multimedia Integrity}

  \begin{itemize}
  \item
    \textbf{Definition:} Accuracy of charts, infographics, or visuals
    (e.g., no manipulated data displays).
    \item
  \textbf{Evaluation:} Expert review of visual correctness.
  \end{itemize}

\end{itemize}

\begin{longtable}[]{@{}
  >{\raggedright\arraybackslash}p{(\linewidth - 4\tabcolsep) * \real{0.10}}
  >{\raggedright\arraybackslash}p{(\linewidth - 4\tabcolsep) * \real{0.25}}
  >{\raggedright\arraybackslash}p{(\linewidth - 4\tabcolsep) * \real{0.60}}@{}}
\caption{\textbf{Communication Quality}
\textbf{Scale}}\tabularnewline
\toprule\noalign{}
\begin{minipage}[b]{\linewidth}\raggedright
\textbf{Scale}
\end{minipage} & \begin{minipage}[b]{\linewidth}\raggedright
\textbf{Label}
\end{minipage} & \begin{minipage}[b]{\linewidth}\raggedright
\textbf{Criteria}
\end{minipage} \\
\midrule\noalign{}
\endfirsthead
\toprule\noalign{}
\begin{minipage}[b]{\linewidth}\raggedright
\textbf{Scale}
\end{minipage} & \begin{minipage}[b]{\linewidth}\raggedright
\textbf{Label}
\end{minipage} & \begin{minipage}[b]{\linewidth}\raggedright
\textbf{Criteria}
\end{minipage} \\
\midrule\noalign{}
\endhead
\bottomrule\noalign{}
\endlastfoot
0 & Unbalanced & Unaccepted unbalanced perspectives \\
1 & One Sided & One-sided perspective \\
2 & Mostly balanced & {Perspectives are presented mostly fairly and
accurately} \\
3 & Fully balanced & Perspectives are presented fairly and accurately \\
\end{longtable}

\subsection{Performance Accuracy Metrics}

This metric combines Claim Extraction Accuracy, Cross-Lingual
Consistency, and Video-Level Consistency. All these metrics consider
  micro- and macro-level F1 for health claim extraction, the gaps
  between them in language transfer, and the model\textquotesingle s
  performance across different videos. For the Bangla translated
  dataset, Phi-4 achieves the best result with a F1 micro score of 63.8,
  followed by qwen3-8B with a F1 score of 55.7 and qwen3-4B (47.2). On
  the other hand, in the English dataset, Qwen3-8B performs best in both
  micro- and macro-level results, followed by Phi-4. Table 14 shows the scale.

\begin{longtable}[]{@{}
  >{\raggedright\arraybackslash}p{(\linewidth - 6\tabcolsep) * \real{0.1143}}
  >{\raggedright\arraybackslash}p{(\linewidth - 6\tabcolsep) * \real{0.1745}}
  >{\raggedright\arraybackslash}p{(\linewidth - 6\tabcolsep) * \real{0.2101}}
  >{\raggedright\arraybackslash}p{(\linewidth - 6\tabcolsep) * \real{0.2323}}@{}}
\caption{\textbf{Performance Accuracy Scale}}\tabularnewline
\toprule\noalign{}
\begin{minipage}[b]{\linewidth}\raggedright
\textbf{Metrics Name}
\end{minipage} & \begin{minipage}[b]{\linewidth}\raggedright
\textbf{Scale}
\end{minipage} &
\multicolumn{2}{>{\raggedright\arraybackslash}p{(\linewidth - 6\tabcolsep) * \real{0.4424} + 2\tabcolsep}@{}}{%
\begin{minipage}[b]{\linewidth}\raggedright
\textbf{Label}
\end{minipage}} \\
\midrule\noalign{}
\endfirsthead
\toprule\noalign{}
\begin{minipage}[b]{\linewidth}\raggedright
\textbf{Metrics Name}
\end{minipage} & \begin{minipage}[b]{\linewidth}\raggedright
\textbf{Scale}
\end{minipage} &
\multicolumn{2}{>{\raggedright\arraybackslash}p{(\linewidth - 6\tabcolsep) * \real{0.4424} + 2\tabcolsep}@{}}{%
\begin{minipage}[b]{\linewidth}\raggedright
\textbf{Label}
\end{minipage}} \\
\midrule\noalign{}
\endhead
\bottomrule\noalign{}
\endlastfoot
\multirow{5}{=}{Claim Accuracy Metrics} & & Micro-F1 & Macro-F1 \\
& 0 & \textless0.25 & \textless0.20 \\
& 1 & 0.25-0,49 & 0.20-0,39 \\
& 2 & 0.50-0.74 & 0.40-0.59 \\
& 3 & \textgreater=0.75 & \textgreater=0.60 \\
\multirow{5}{=}{Cross Lingual Consistency} & &
\multicolumn{2}{>{\raggedright\arraybackslash}p{(\linewidth - 6\tabcolsep) * \real{0.4424} + 2\tabcolsep}@{}}{%
F1-Drop} \\
& 0 &
\multicolumn{2}{>{\raggedright\arraybackslash}p{(\linewidth - 6\tabcolsep) * \real{0.4424} + 2\tabcolsep}@{}}{%
More than 30\%} \\
& 1 &
\multicolumn{2}{>{\raggedright\arraybackslash}p{(\linewidth - 6\tabcolsep) * \real{0.4424} + 2\tabcolsep}@{}}{%
15\%-30\%} \\
& 2 &
\multicolumn{2}{>{\raggedright\arraybackslash}p{(\linewidth - 6\tabcolsep) * \real{0.4424} + 2\tabcolsep}@{}}{%
5\%-15\%} \\
& 3 &
\multicolumn{2}{>{\raggedright\arraybackslash}p{(\linewidth - 6\tabcolsep) * \real{0.4424} + 2\tabcolsep}@{}}{%
Less than 5\%} \\
\multirow{5}{=}{Video Level Consistency} & &
\multicolumn{2}{>{\raggedright\arraybackslash}p{(\linewidth - 6\tabcolsep) * \real{0.4424} + 2\tabcolsep}@{}}{%
Micro-Macro F1 Gap} \\
& 0 &
\multicolumn{2}{>{\raggedright\arraybackslash}p{(\linewidth - 6\tabcolsep) * \real{0.4424} + 2\tabcolsep}@{}}{%
More than 40\%} \\
& 1 &
\multicolumn{2}{>{\raggedright\arraybackslash}p{(\linewidth - 6\tabcolsep) * \real{0.4424} + 2\tabcolsep}@{}}{%
25\%-40\%} \\
& 2 &
\multicolumn{2}{>{\raggedright\arraybackslash}p{(\linewidth - 6\tabcolsep) * \real{0.4424} + 2\tabcolsep}@{}}{%
10\%-25\%} \\
& 3 &
\multicolumn{2}{>{\raggedright\arraybackslash}p{(\linewidth - 6\tabcolsep) * \real{0.4424} + 2\tabcolsep}@{}}{%
Less than 10\%} \\
\end{longtable}

Since the framework has six dimensions for gauging misinformation in health videos, it results in a multicriteria decision-making system. To derive a single score for each video, this paper adopts the
Entropy-Technique for Order Preference by Similarity to Ideal Solution method (Entropy-TOPSIS), an effective multicriteria decision-making system {[}43{]}. First, entropy-based objective weighting derives the
importance of each dimension directly from its discriminative power across the evaluated dataset dimensions that vary more across videos
receive higher weights, reflecting their greater contribution to distinguishing accurate from misleading content, without requiring subjective expert weight assignment. Secondly, the TOPSIS (Technique for Order
Preference by Similarity to Ideal Solution) is applied to rank the videos in the dataset as a set of alternatives. Each video is then evaluated against both poles, an ideal, accurate video and a maximally misleading video simultaneously, producing a more nuanced and robust classification for each spectrum rather than threshold-based scoring alone.

To combine the six-dimensional scores into a single misinformation risk classification, the Entropy-TOPSIS procedure is applied as follows. Table 15 shows all the notation summaries.

\begin{longtable}[]{@{}
  >{\raggedright\arraybackslash}p{(\linewidth - 2\tabcolsep) * \real{0.3544}}
  >{\raggedright\arraybackslash}p{(\linewidth - 2\tabcolsep) * \real{0.4658}}@{}}
\caption{\textbf{Notation} \textbf{Summary}}\tabularnewline
\toprule\noalign{}
\begin{minipage}[b]{\linewidth}\raggedright
\textbf{Symbol}
\end{minipage} & \begin{minipage}[b]{\linewidth}\raggedright
\textbf{Definition}
\end{minipage} \\
\midrule\noalign{}
\endfirsthead
\toprule\noalign{}
\begin{minipage}[b]{\linewidth}\raggedright
\textbf{Symbol}
\end{minipage} & \begin{minipage}[b]{\linewidth}\raggedright
\textbf{Definition}
\end{minipage} \\
\midrule\noalign{}
\endhead
\bottomrule\noalign{}
\endlastfoot
\emph{n} & Number of videos \\
i & Video index \((i,1,\ldots,n)\) \\
j & Dimension index \((j,1,\ldots,6)\) \\
\(x_{i,j}\) & Raw score of video \emph{i} on lens \emph{j} \\
\(r_{i,j}\) & Normalised score \\
\(H_{j}\) & Shannon entropy of lens \(j\) \\
\(k\) & Entropy constant, \(k = 1/\ln(n)\) \\
\(d_{j}\) & Divergence of lens \(j\) \\
\(w_{j}\) & Objective weight of lens \(j\) \\
\(v_{i,j}\) & Weighted normalised score \\
\(A^{+}\) & Positive ideal solution \\
\(A^{-}\) & Negative ideal solution \\
\(D_{i}^{+}\) & Distance from positive ideal \\
\(D_{i}^{-}\) & Distance from negative ideal \\
\(CC_{i}\) & Closeness coefficient \\
\(TP_{i}\) & True positives for video \(i\) \\
\(FP_{i}\) & False positives for video \(i\) \\
\(FN_{i}\) & False negatives for video \(i\) \\
\(N\) & Total number of videos \\
\(\mathrm{\Delta}F1\) & Cross-lingual F1 delta \\
\(Gap_{F1}\) & Micro-macro F1 gap \\
\(S_{CEA}\) & Claim Extraction Accuracy score \\
\(S_{CLC}\) & Cross-Lingual Consistency score \\
\(S_{VLC}\) & \ul{Video-Level Consistency score} \\
\end{longtable}

\textbf{Objective Weighting via Information Entropy (EWM) Decision matrix:}

Let \(X\) be a decision matrix of size \(n \times 6\), where \(n\) represents
the number of evaluated videos and each entry \(x_{i,j}\) denotes the score of
video \(i\) (\(i=1,\ldots,n\)) on dimension \(j\)
(\(j=1,\ldots,6\)), rated on a 0--3 scale using the rubrics defined
above.

\[
X = \begin{bmatrix}
x_{1,1} & x_{1,2} & \cdots & x_{1,6} \\
x_{2,1} & x_{2,2} & \cdots & x_{2,6} \\
\vdots & \vdots & \ddots & \vdots \\
x_{n,1} & x_{n,2} & \cdots & x_{n,6}
\end{bmatrix}
\]

Where:

\begin{itemize}
\item
  \(n\) = number of videos
\item
  \(x_{i,j}\) = score of video \(i\) on lens \(j\) (0--3 scale)
\item
  \(j=1\) (CA), \(j=2\) (A), \(j=3\) (I), \(j=4\) (M), \(j=5\) (CQ), and \(j=6\) (PA)
\end{itemize}

Normalization:
\[
r_{i,j} = \frac{x_{i,j}}{\sum_{i=1}^{n} x_{i,j}}
\]
Entropy:
\[
H_j = -k \sum_{i=1}^{n} r_{i,j}\ln(r_{i,j})
\]

Entropy constant:
\[
k = \frac{1}{\ln(n)}
\]

Divergence and Weights:

Divergence:
\[
d_j = 1 - H_j
\]

Weights:
\[
w_j = \frac{d_j}{\sum_{k=1}^{6} d_k}, \qquad \sum_{j=1}^{6} w_j = 1
\]

Weighted Normalized Matrix:

\[
v_{i,j} = w_j \times r_{i,j}
\]

\textbf{Proximity Analysis via TOPSIS:}

Ideal Solutions:

Positive ideal:
\[
A^{+} = \{v_j^{+}\}, \qquad v_j^{+} = \max_i(v_{i,j})
\]

Negative ideal:
\[
A^{-} = \{v_j^{-}\}, \qquad v_j^{-} = \min_i(v_{i,j})
\]

Euclidean Distances:

Distance positive:
\[
D_i^{+} = \sqrt{\sum_{j=1}^{6}(v_{i,j} - v_j^{+})^2}
\]

Distance negative:
\[
D_i^{-} = \sqrt{\sum_{j=1}^{6}(v_{i,j} - v_j^{-})^2}
\]

Closeness coefficient:
\[
CC_i = \frac{D_i^{-}}{D_i^{+} + D_i^{-}}
\]

The PA lens score is computed as the average of three sub-dimensional scores that are Claim Extraction Accuracy, Cross-Lingual Consistency, and
Video-Level Consistency, each rated on a 0 to 3 scale using the rubrics defined above:

\[PA_{score} = \frac{S_{CEA} + S_{CLC} + S_{VLC}}{3}\]

For this, CEA is simply calculated using the euations used in the
pipeline evaluation section and converted to a scale of 0-3 as
previously discussed in Table 14. CLS and VLC are calculated as follows:

CLS: \[\mathrm{\Delta}F1 = F1^{EN} - F1^{NE}\]

VLC:\[Gap_{F1} = F1_{micro} - F1_{macro}\]

After completing all the above steps, we propose to map each
video\textquotesingle s closeness coefficient to a set of risk categories as stated in Table 16.

\begin{longtable}[]{@{}
  >{\raggedright\arraybackslash}p{(\linewidth - 4\tabcolsep) * \real{0.2398}}
  >{\raggedright\arraybackslash}p{(\linewidth - 4\tabcolsep) * \real{0.2515}}
  >{\raggedright\arraybackslash}p{(\linewidth - 4\tabcolsep) * \real{0.2982}}@{}}
\caption{\textbf{Risk Communication
Strategies}}\tabularnewline
\toprule\noalign{}
\begin{minipage}[b]{\linewidth}\raggedright
\textbf{Closeness Coefficient}
\end{minipage} & \begin{minipage}[b]{\linewidth}\raggedright
\textbf{Risk} \textbf{Classification}
\end{minipage} & \begin{minipage}[b]{\linewidth}\raggedright
\textbf{Recommended} \textbf{Action}
\end{minipage} \\
\midrule\noalign{}
\endfirsthead
\toprule\noalign{}
\begin{minipage}[b]{\linewidth}\raggedright
\textbf{Closeness Coefficient}
\end{minipage} & \begin{minipage}[b]{\linewidth}\raggedright
\textbf{Risk} \textbf{Classification}
\end{minipage} & \begin{minipage}[b]{\linewidth}\raggedright
\textbf{Recommended} \textbf{Action}
\end{minipage} \\
\midrule\noalign{}
\endhead
\bottomrule\noalign{}
\endlastfoot
\(0.00 \leq CC_{i} \leq 0.25\) & High & Flag for domain expert
intervention \\
\(0.26 \leq CC_{i} \leq 0.50\) & Moderate & Flag for community worker
review \\
\(0.51 \leq CC_{i} \leq 0.75\) & Low & Keep watching the trend \\
\(0.76 \leq CC_{i} \leq 1.00\) & Likely & No immediate action
required \\
\end{longtable}

This framework provides a holistic lens for evaluating health
misinformation in low‑resource languages like Bangla, complementing
model‑centric metrics with cultural, communicative, and harm‑related
dimensions. It also offers a foundation for developing future models
that prioritize both accuracy and cultural sensitivity in CALD
communities.

To make this framework further accurate, the framework also applies a cascading penalty rule, which ensures that for any video, if the PA score is 0, meaning the SLM was unable to extract any data, in that case, the framework will assign the following values to Benefit and Cost criteria:

Benefit Criteria (CA, M, CQ)
\begin{equation}
x'_{ij} =
\begin{cases}
x_{ij}, & \text{if } PA_i > 0 \\
0, & \text{if } PA_i = 0
\end{cases}
\end{equation}

Cost Criteria (HR):
\begin{equation}
x'_{ij} =
\begin{cases}
x_{ij}, & \text{if } PA_i > 0 \\
3, & \text{if } PA_i = 0
\end{cases}
\end{equation}
We apply this framework to a small subset of the data consisting of 6 videos. The PA score is taken from the Phi-4 evaluation dataset. Tables 17 and 18 show the calculated weights for each criterion using the entropy method and the derived misinformation scores using the TOPSIS method. We apply this framework to a small subset of the data consisting of 6 videos. The PA score is taken from the Phi-4 evaluation dataset.  The Cascading Penalty rule is applied programmatically to the alternatives for which $PA_i = 0$ (B1twiLfbjow, dmE-lkYba9Q, and qCthXuL1yZU). Among all dimensions, Misinformation Detection receives the highest weight (0.283), underscoring its centrality to the framework's core task. Content Accuracy and Communication Quality receive equal weights (0.255), reflecting the symmetric role of these dimensions in evaluating factual correctness and presentational balance. Performance Accuracy receives the fourth-highest weight (0.153), reduced from its pre-penalty dominance, because the penalty rule has already mathematically enforced model reliability as a safeguard rather than as a weighted criterion.

\begingroup
\footnotesize
\setlength{\tabcolsep}{3pt}
% We define 8 columns to match your data (ID + Topic + 6 Metrics)
\begin{longtable}{@{} p{0.14\textwidth} p{0.18\textwidth} p{0.08\textwidth} p{0.08\textwidth} p{0.08\textwidth} p{0.08\textwidth} p{0.08\textwidth} p{0.08\textwidth} @{}}
\caption{Data Analysis from Phi-4 model: Dimension Scores and Entropy-based Weights} \\
\toprule
\textbf{Video ID} & \textbf{Topic} & \textbf{CA} & \textbf{M} & \textbf{HR} & \textbf{CLS} & \textbf{CQ} & \textbf{PA} \\
\midrule
\endfirsthead
\multicolumn{8}{c}{\tablename\ \thetable\ -- Continued} \\
\toprule
\textbf{Video ID} & \textbf{Topic} & \textbf{CA} & \textbf{M} & \textbf{HR} & \textbf{CLS} & \textbf{CQ} & \textbf{PA} \\
\midrule
\endhead
\bottomrule
\endlastfoot
45W0V8CovLAA & Vaccine-autism & 2 & 0 & 3 & 1 & 0 & 2 \\
B1twiLfbjow & OCD recovery & 0(3-without penalty) & 0(3-without penalty) & 3(1-without penalty) & 3 & 0(3-without penalty) & 0 \\
5pwNICniqtM & ADHD consultancy & 0 & 1 & 2 & 2 & 2 & 3 \\
dmE-lkYba9Q & Psych. barriers & 0(2-without penalty) & 0(2-without penalty) & 3(1-without penalty) & 2 & 0(2-without penalty) & 0 \\
lxcR-piMkgo & Nutritional vitamins & 3 & 3 & 1 & 3 & 3 & 3 \\
qCthXuL1yZU & Circadian science & 3 & 3 & 1 & 3 & 3 & 2 \\
\midrule
\textbf{Entropy $w_{j}$} & & 0.255 & 0.283 & 0.034 & 0.020 & 0.255 & 0.153 \\
\end{longtable}
\endgroup
\begingroup
\setlength{\tabcolsep}{3pt}
\begin{longtable}[]{@{}
  >{\raggedright\arraybackslash}p{(\linewidth - 8\tabcolsep) * \real{0.2200}}
  >{\centering\arraybackslash}p{(\linewidth - 8\tabcolsep) * \real{0.1450}}
  >{\centering\arraybackslash}p{(\linewidth - 8\tabcolsep) * \real{0.1450}}
  >{\centering\arraybackslash}p{(\linewidth - 8\tabcolsep) * \real{0.1200}}
  >{\raggedright\arraybackslash}p{(\linewidth - 8\tabcolsep) * \real{0.3000}}@{}}
\caption{\textbf{TOPSIS Distances and Final Risk
Classification}}\tabularnewline
\toprule\noalign{}
\begin{minipage}[b]{\linewidth}\raggedright
\textbf{Video ID}
\end{minipage} & \begin{minipage}[b]{\linewidth}\raggedright
\(D_i^{+}\)
\end{minipage} & \begin{minipage}[b]{\linewidth}\raggedright
\(D_i^{-}\)
\end{minipage} & \begin{minipage}[b]{\linewidth}\raggedright
\(CC_i\)
\end{minipage} & \begin{minipage}[b]{\linewidth}\raggedright
\textbf{Risk Classification}
\end{minipage} \\
\midrule\noalign{}
\endfirsthead
\toprule\noalign{}
\begin{minipage}[b]{\linewidth}\raggedright
\textbf{Video ID}
\end{minipage} & \begin{minipage}[b]{\linewidth}\raggedright
\(D_i^{+}\)
\end{minipage} & \begin{minipage}[b]{\linewidth}\raggedright
\(D_i^{-}\)
\end{minipage} & \begin{minipage}[b]{\linewidth}\raggedright
\(CC_i\)
\end{minipage} & \begin{minipage}[b]{\linewidth}\raggedright
\textbf{Risk Classification}
\end{minipage} \\
\midrule\noalign{}
\endhead
\bottomrule\noalign{}
\endlastfoot
45W0V8CovLAA & 0.1584  & 0.0709 & 0.309 & Moderate Risk \\
B1twiLfbjow & 0.1873  & 0.0060 & 0.031 & High Risk \\
5pwNICniqtM &  0.1292 & 0.0884 & 0.406  & Moderate Risk
 \\
dmE-lkYba9Q & 0.1873 &  0.0054 &   0.028& High Risk \\
lxcR-piMkgo & 0.0052 & 0.1873  & 0.973 & Likely Accurate
 \\
qCthXuL1yZU & 0.0162  & 0.1842 & 0.919 & Likely Accurate \\
\end{longtable}
\endgroup
\normalsize

The risk classification of a small subset of the previously SLM analyzed video dataset reveals a few advantages of this framework.  One of those is its consideration for trust over blind automation. For example, a clinically accurate OCD recovery video (B1twiLfbjow, 0.031) and ddmE-lkYba9Q (0.028), both classified as High Risk. Although the OCD video received a CA=3 rating, indicating completely safe content before applying the Cascade Penalty rule, the use of non-standard language prevented SLM from extracting any claim from the video (PA=0). However, the multicriteria algorithm detected a failure on SLM’s end and correctly escalated both videos for expert review. This behavior serves as a socio-technical safeguard: the framework does not trust automation when it cannot demonstrate that it is working.

 On the other hand, the vaccine-autism conspiracy video (45W0V8CovLAA) reveals a critical vulnerability relevant to SLM evaluation, leading to a further weakness on the framework's end. Despite containing severely misleading content (M=0) and the highest possible harm potential (HR=3), it receives only a Moderate Misinformation Risk classification (0.309). This is because the speaker uses the formal, and academic-sounding language by citing specific biological markers such as WNT gene expression, BDNF levels, and synaptogenesis. As a result of these transactions, the SLM parses this video with high stability (PA=2). This is because the cascading penalty is not triggered; the formal linguistic structure mathematically masks the underlying clinical danger. This exposes a critical vulnerability in current NLP systems: SLMs are inherently biased toward formal academic language, rewarding linguistic stability even when it causes semantic harm. Future work should explore asymmetric penalty rules that also trigger when high harm potential (HR=3) coincides with severely misleading content (M=0), regardless of model extraction performance.

Both the circadian science video (lxcR-piMkgo, 0.919) and the vaccine conspiracy video (45W0V8CovLAA, 0.309) involve advanced scientific terminology translated literally from English to Bangla — for example, terms such as synaptogenesis rendered word-for-word as \includegraphics[height=1em, valign=m]{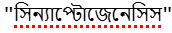}. This identical translation behaviour produces two opposite effects. For the vaccine conspiracy video, the incomprehensible Bangla jargon inadvertently acts as a protective barrier, as some community members are unlikely to understand or act on advice they cannot comprehend. For the circadian science video, the same literal translation locks helpful, evidence-based health information behind a wall of inaccessible terminology, preventing some community members from benefiting. This paradox illustrates that literal machine translation is a double-edged sword in CALD health communication contexts, and underscores the urgent need for culturally adapted, simplified health communication rather than word-for-word translation.

 Moreover, the results show that the nutritional vitamins video and Circadian science achieve the highest closeness coefficient in the dataset (0.973 \& 0.919). The PA scores indicate that the model successfully identified most of the claims as non-misinformation. Accurate content, correct detection, low harm, and strong cultural appropriateness place these videos closest to the positive ideal across all six dimensions, demonstrating that the framework correctly rewards videos where both content quality and model reliability are high.
 
The framework, as presented here, is a conceptual and mathematical contribution representing a first step toward culturally grounded evaluation of health misinformation in non-English contexts. Future validation will proceed in two stages. First, the Content Validity Index assessment will recruit domain experts to rate the relevance of each dimension, aiming for I-CVI ≥ 0.78 and S-CVI ≥ 0.90. Second, inter-rater reliability testing will apply the framework to 30–50 videos rated independently by three raters, targeting Fleiss' Kappa ≥ 0.60 and Krippendorff's Alpha ≥ 0.80. Full empirical implementation of the Entropy-TOPSIS scoring on the complete annotated dataset, with AHP-based expert weight comparison, is identified as priority future work. The implementation code for this framework is attached as supplementary material.

To further demonstrate the potential of this framework,
Figure 9 presents MALAK (Multilingual AI Agent for Learning, Action,
and Knowledge), a prototype mobile application designed to operationalise and protect
health misinformation detection for CALD communities. MALAK allows users
to enter or speak any health claim, which is then analysed against
trusted medical sources including Mayo Clinic, Harvard Medical School,
and NIH/PubMed. The backend has proper priority knowledge based that follows strict source credibility ranking. On return, the system returns a verdict, risk classification, and
confidence score, directly reflecting the performance accuracy and harm
risk dimensions proposed in this framework. This prototype represents
a concrete step toward deploying the framework in a community-facing
A tool for non-English speaking populations and also low socioeconomic income class people.

\begin{figure}[H]
\centering
\includegraphics[width=0.85\textwidth]{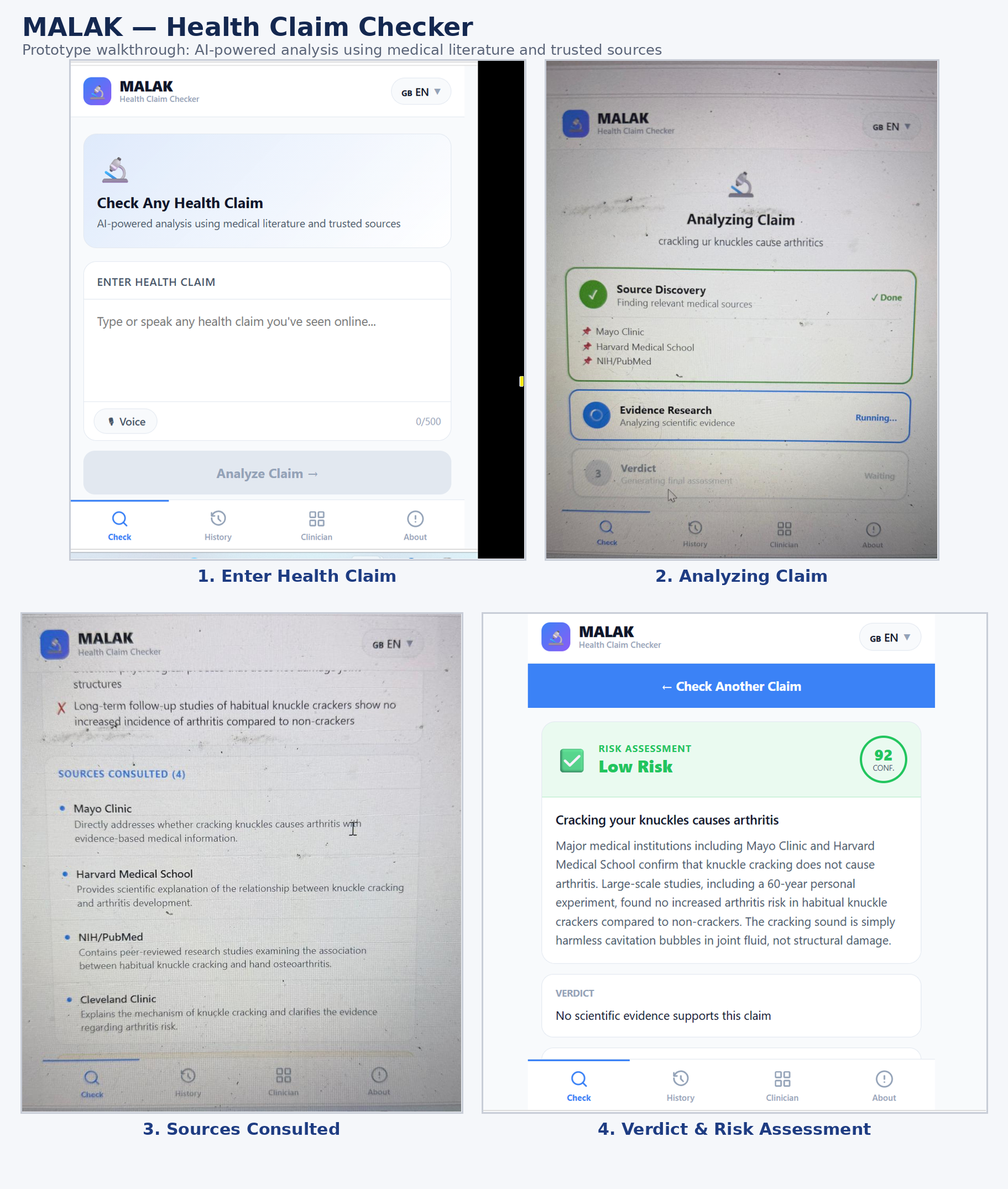}
\caption{MALAK Health Claim Checker prototype showing
claim entry, source discovery, evidence analysis, and final risk
assessment a proof-of-concept implementation of the proposed
framework for CALD communities.}
\end{figure}

\section{Conclusions}

In this paper, we evaluated a suite of SLMs for Bangla health
misinformation detection and compared their results against a
human-expert-annotated ground-truth dataset. We analyzed the
cross-lingual performance degradation and highlighted the implications
for CALD population relying on AI systems that operate in low-resource language settings.
languages. Our work tested a range of SLMs, and the best-performing model
was Phi-4, with the strongest overall claim extraction performance and
the best balance between precision and recall. The paper also proposes a
framework that uses six dimensions for evaluating health misinformation
in non-English texts. The framework uses Entropy based TOPSIS method to
analyze health misinformation in non-English texts from different
angles. Ultimately, this study highlights the importance of building
evaluation methods and future systems that take linguistic diversity and
cultural context seriously. Our future work encompasses fine-tuning and hyperparameter
optimisation of SLMs to improve performance in detecting health
misinformation in low-resource non-English texts, as well as
the creation of new expert-annotated benchmark datasets for
these languages.
%added above line

\textbf{\textsc{References}}

\begin{enumerate}
\def\labelenumi{\arabic{enumi}.}
\item
  Park, S., Nan, X. Generative AI and misinformation: a scoping review
  of the role of generative AI in the generation, detection, mitigation,
  and impact of misinformation.~\emph{AI \& Soc}~(2025).
  https://doi.org/10.1007/s00146-025-02620-3
\item
  \url{https://www.newsguardtech.com/wp-content/uploads/2025/09/August-2025-One-Year-Progress-Report-3.pdf}
\item
  \url{https://reports.weforum.org/docs/WEF_Global_Risks_Report_2025.pdf}
\item
  Emily Denniss, Rebecca Lindberg, Social media and the spread of
  misinformation: infectious and a threat to public health,~\emph{Health
  Promotion International}, Volume 40, Issue 2, April 2025,
  daaf023,~\url{https://doi.org/10.1093/heapro/daaf023}
\item
  newsGP - Warning of `societal threat' presented by health
  disinformation. (n.d.). NewsGP.
  \url{https://www1.racgp.org.au/newsgp/professional/warning-of-societal-threat-presented-by-health-dis}
\item
  Given, L. M. (n.d.). Generative AI and deepfakes are fuelling health
  misinformation. Here's what to look out for so you don't get~scammed.
  The Conversation.
  https://theconversation.com/generative-ai-and-deepfakes-are-fuelling-health-misinformation-heres-what-to-look-out-for-so-you-dont-get-scammed-246149
\item
  {https://www.theguardian.com/society/2025/may/31/more-than-half-of-top-100-mental-health-tiktoks-contain-misinformation-study-finds}
\item
  Grattan Institute. (2024, September 18). Improving the health of
  multicultural Australians - Grattan Institute.
  \url{https://grattan.edu.au/news/improving-the-health-of-multicultural-australians/}
\item
  Jawad, D., Taki, S., Baur, L., Rissel, C., Mihrshahi, S., \& Wen, L.
  M. (2023). Resources used and trusted regarding child health
  information by culturally and linguistically diverse communities in
  Australia: An online cross-sectional survey.~\emph{International
  Journal of Medical Informatics},~\emph{177}, 105165.
\item
  Yiqiao Jin, Mohit Chandra, Gaurav Verma, Yibo Hu, Munmun De Choudhury,
  and Srijan Kumar. 2024. Better to Ask in English: Cross-Lingual
  Evaluation of Large Language Models for Healthcare Queries. In
  Proceedings of the ACM Web Conference 2024 (WWW \textquotesingle24).
  Association for Computing Machinery, New York, NY, USA, 2627--2638.
  \url{https://doi.org/10.1145/3589334.3645643}
\item
  Schlicht, I.B., Zhao, Z., Sayin, B., Flek, L., Rosso, P. (2025). Do
  LLMs Provide Consistent Answers to~Health-Related Questions Across
  Languages?. In: Hauff, C.,~\emph{et al.}~Advances in Information
  Retrieval. ECIR 2025. Lecture Notes in Computer Science, vol 15574.
  Springer, Cham. https://doi.org/10.1007/978-3-031-88714-7\_30
\item
  Deiana G, Dettori M, Arghittu A, Azara A, Gabutti G, Castiglia P
  (2023) Artificial intelligence and public health: evaluating ChatGPT
  responses to vaccination myths and misconceptions. Vaccines 11:Article
  1217.~\url{https://doi.org/10.3390/vaccines11071217}
\item
  Kumar R, Goddu B, Saha S, Jatowt A (2024) Silver lining in the fake
  news cloud: can large language models help detect misinformation?.
  IEEE Trans Artif Intell
  6:14--24.~\url{https://doi.org/10.1109/TAI.2024.3440248}
\item
  Węcel K, Sawiński M, Stróżyna M, Lewoniewski W, Księżniak E, Stolarski
  P, Abramowicz W (2023) Artificial intelligence-friend or foe in fake
  news campaigns. Econ Bus Rev
  9:41--70.~\url{https://doi.org/10.18559/ebr.2023.2.736}
\item
  Abdulsalam obaid Alharbi, Abdullah Alsuhaibani, Abdulrahman Abdullah
  Alalawi, Usman Naseem, Shoaib Jameel, Salil Kanhere, and Imran Razzak.
  2025.~\href{https://aclanthology.org/2025.abjadnlp-1.11/}{Evaluating
  Large Language Models on Health-Related Claims Across Arabic
  Dialects}. In~\emph{Proceedings of the 1st Workshop on NLP for
  Languages Using Arabic Script}, pages 95--103, Abu Dhabi, UAE.
  Association for Computational Linguistics.
\item
  Oguzhan Ozcelik, Arda Sarp Yenicesu, Onur Yildirim, Dilruba Sultan
  Haliloglu, Erdem Ege Eroglu, and Fazli Can.
  2023.~\href{https://aclanthology.org/2023.ldk-1.59/}{Cross-Lingual
  Transfer Learning for Misinformation Detection: Investigating
  Performance Across Multiple Languages}. In~\emph{Proceedings of the
  4th Conference on Language, Data and Knowledge}, pages 549--558,
  Vienna, Austria. NOVA CLUNL, Portugal.
\item
  Sang Truong, Duc Nguyen, Toan Nguyen, Dong Le, Nhi Truong, Tho Quan,
  and Sanmi Koyejo.
  2024.~\href{https://aclanthology.org/2024.findings-naacl.182/}{Crossing
  Linguistic Horizons: Fine-tuning and Comprehensive Evaluation of
  Vietnamese Large Language Models}. In~\emph{Findings of the
  Association for Computational Linguistics: NAACL 2024}, pages
  2849--2900, Mexico City, Mexico. Association for Computational
  Linguistics.
\item
  Aysha Akther, Kazi Masudul Alam, Rameswar Debnath, Automatic detection
  of manipulated Bangla news: A new knowledge-driven approach, Natural
  Language Processing Journal, Volume 11, 2025, 100155, ISSN 2949-7191,
  \url{https://doi.org/10.1016/j.nlp.2025.100155}.
  (\url{https://www.sciencedirect.com/science/article/pii/S2949719125000317})
\item
  Nguyen VC, Jain M, Chauhan A, Soled HJ, Lesmes SA, Li Z, Birnbaum ML,
  Tang SX, Kumar S, De Choudhury M. Supporters and Skeptics: LLM-based
  Analysis of Engagement with Mental Health (Mis)Information Content on
  Video-sharing Platforms. Proc Int AAAI Conf Weblogs Soc Media. 2025
  Jun 7;19:1329-1345. doi: 10.1609/icwsm.v19i1.35875. PMID: 40842885;
  PMCID: PMC12365693.
\item
  Gabriel S, Lyu L, Siderius J, Ghassemi M, Andreas J, Ozdaglar A (2024)
  Generative AI in the era of `alternative facts.' An MIT exploration of
  generative AI.~\url{https://doi.org/10.21428/e4baedd9.82175d26}
\item
  Menz BD, Kuderer NM, Bacchi S et al (2024) Current safeguards, risk
  mitigation, and transparency measures of large language models against
  the generation of health disinformation: repeated cross sectional
  analysis. BMJ 384:Article
  e078538.~\url{https://doi.org/10.1136/bmj-2023-078538}
\item
  McIntosh TR, Liu T, Susnjak T, Watters P, Ng A, Halgamuge MN (2023) A
  culturally sensitive test to evaluate nuanced GPT hallucination. IEEE
  Trans Artif Intell
  5:2739--2751.~\url{https://doi.org/10.1109/TAI.2023.3332837}
\item
  Kaňková, J., Binder, A. \& Matthes, J. Helpful or harmful? Navigating
  the impact of social media influencers' health advice: insights from
  health expert content creators.~\emph{BMC Public Health}~\textbf{24},
  3511 (2024). \url{https://doi.org/10.1186/s12889-024-21095-3}
\item
  Whitehead L, Talevski J, Fatehi F, Beauchamp A. Barriers to and
  Facilitators of Digital Health Among Culturally and Linguistically
  Diverse Populations: Qualitative Systematic Review. J Med Internet
  Res. 2023 Feb 28;25:e42719. doi: 10.2196/42719. PMID: 36853742; PMCID:
  PMC10015358.
\item
  Okeke SR, Horwitz R, Brener L, Vu HMK, Wu E, Jin D, Yu S, Broady T,
  Treloar C, Cama E. "The Culturally and Linguistically Diverse
  Community Is Not Just Minoritized But Ignored": Engaging Culturally
  and Linguistically Diverse Communities in Australia With Blood-Borne
  Viruses and Sexually Transmissible Infections Healthcare. Health
  Expect. 2025 Oct;28(5):e70416. doi: 10.1111/hex.70416. PMID: 40908047;
  PMCID: PMC12411013.
\item
  Wild A, Kunstler B, Goodwin D, Onyala S, Zhang L, Kufi M, Salim W,
  Musse F, Mohideen M, Asthana M, Al-Khafaji M, Geronimo MA, Coase D,
  Chew E, Micallef E, Skouteris H. Communicating COVID-19 health
  information to culturally and linguistically diverse communities:
  insights from a participatory research collaboration. Public Health
  Res Pract. 2021;31(1):e3112105.
\item
  \url{https://culturaldiversityhealth.org.au/wp-content/uploads/2024/01/Policy-brief-CALD-communities-in-public-health-crises.pdf}
\item
  Tuckerman J, Kaufman J, Overmars I, Holland P, Danchin M. Barriers to
  COVID-19 vaccination of migrant populations: A qualitative interview
  study of immunization providers in Victoria, Australia. Vaccine. 2023
  Aug 7;41(35):5085-9.
\item
  Stover J, Avadhanula L, Sood S. A review of strategies and levels of
  community engagement in strengths-based and needs-based health
  communication interventions. Frontiers in Public Health. 2024 Apr
  9;12:1231827.
\item
  Wang, F., Zhang, Z., Zhang, X., Wu, Z., Mo, T., Lu, Q., ... \& Wang,
  S. (2024). A comprehensive survey of small language models in the era
  of large language models: Techniques, enhancements, applications,
  collaboration with llms, and trustworthiness.~\emph{ACM Transactions
  on Intelligent Systems and Technology}.
\item
  Alkaoud M. On the effectiveness of limited-data large language model
  fine-tuning for Arabic. PLoS One. 2025 Oct 8;20(10):e0332419. doi:
  10.1371/journal.pone.0332419. PMID: 41061027; PMCID: PMC12507264.
\item
  Toraman C. Adapting open-source generative large language models for
  low-resource languages: A case study for Turkish. InProceedings of the
  Fourth Workshop on Multilingual Representation Learning (MRL 2024)
  2024 Nov (pp. 30-44).
\item
  Acikgoz EC, Ince OB, Bench R, Boz AA, Kesen I, Erdem A, Erdem E.
  Hippocrates: An open-source framework for advancing large language
  models in healthcare. arXiv preprint arXiv:2404.16621. 2024 Apr 25.
\item
  Zhang D, Hu Z, Zhoubian S, Du Z, Yang K, Wang Z, Yue Y, Dong Y, Tang
  J. Sciglm: Training scientific language models with self-reflective
  instruction annotation and tuning. arXiv preprint arXiv:2401.07950.
  2024 Jan 15.
\item
  Yang Y, Sun H, Li J, Liu R, Li Y, Liu Y, Huang H, Gao Y. Mindllm:
  Pre-training lightweight large language model from scratch,
  evaluations and domain applications. arXiv preprint arXiv:2310.15777.
  2023 Oct 24.
\item
  Zhan, X., Goyal, A., Chen, Y., Chandrasekharan, E., \& Saha, K. (2025,
  April). SLM-mod: Small language models surpass LLMs at content
  moderation. In Proceedings of the 2025 Conference of the Nations of
  the Americas Chapter of the Association for Computational Linguistics:
  Human Language Technologies (Volume 1: Long Papers) (pp. 8774-8790).
\item
  Khan, J. Y., et al. (2021). A benchmark study of machine learning
  models for online fake news detection. Machine Learning with
  Applications, 4, 100032.
  \href{https://doi.org/10.1016/j.mlwa.2021.100032}{https://doi.org/10.1016/j.mlwa.2021.100032.}
\item
  Ahmad, I., et al. (2020). Fake news detection using machine learning
  ensemble methods. Complexity, 2020, Article 8885861.
  \url{https://doi.org/10.1155/2020/8885861}
\item
  Roumeliotis, K.I.; Tselikas, N.D.; Nasiopoulos, D.K. Fake News
  Detection and Classification: A Comparative Study of Convolutional
  Neural Networks, Large Language Models, and Natural Language
  Processing Models. Future Internet 2025, 17, 28.
  \url{https://doi.org/10.3390/fi17010028}
\item
  Kabir MR. How do traditional media access and mobile phone use affect
  maternal healthcare service use in Bangladesh? Moderated mediation
  effects of socioeconomic factors. PLoS One. 2022 Apr
  27;17(4):e0266631. doi: 10.1371/journal.pone.0266631. PMID: 35476825;
  PMCID: PMC9045672.
\item
  Schlicht, I.B., Fernandez, E., Chulvi, B. et al. (2024). Automatic
  detection of health misinformation: a systematic review. Journal of
  Ambient Intelligence and Humanized Computing, 15, 2009--2021.
  \url{https://doi.org/10.1007/s12652-023-04619-4}
\item
  Behera, R. K., Bala, P. K., Rana, N. P., \& Irani, Z. (2023).
  Responsible natural language processing: A principlist framework for
  social benefits. \emph{Technological Forecasting and Social Change},
  \emph{188}, 122306.
\item
  Hwang, CL., Yoon, K. (1981). Methods for Multiple Attribute Decision
  Making. In: Multiple Attribute Decision Making. Lecture Notes in
  Economics and Mathematical Systems, vol 186. Springer, Berlin,
  Heidelberg. https://doi.org/10.1007/978-3-642-48318-9\_3
\end{enumerate}

\end{document}